\documentclass[runningheads]{llncs}

\usepackage{eccv}

\usepackage{eccvabbrv}

\usepackage{graphicx}
\usepackage{booktabs}
\usepackage{xcolor,colortbl}
\usepackage{multirow}
\usepackage{amsmath}

\usepackage{color}
\definecolor{crimson}{rgb}{0.86, 0.08, 0.24}
\definecolor{gray}{rgb}{0.5,0.5,0.5}
\definecolor{green}{rgb}{0, 0.4, 0}
\definecolor{mahogany}{rgb}{0.75, 0.25, 0.0}
\definecolor{purple}{rgb}{0.6, 0, 0.6}
\definecolor{darkgreen}{rgb}{0, 0.4, 0}
\definecolor{frenchblue}{rgb}{0.0, 0.45, 0.73}
\definecolor{magenta}{rgb}{1,0,1}
\definecolor{pink}{rgb}{1,0.412,0.706}
\definecolor{forestgreen}{RGB}{59, 192, 0}
\definecolor{goldenyellow}{RGB}{255, 192, 0}

\definecolor{lightyellow}{rgb}{1,1, 0.6}
\definecolor{lightorange}{rgb}{1, 0.8, 0.6}
\definecolor{lightred}{rgb}{1, 0.6, 0.6}

\newcommand{\redcell}[0]{\cellcolor{lightred}}
\newcommand{\orgcell}[0]{\cellcolor{lightorange}}
\newcommand{\ylwcell}[0]{\cellcolor{lightyellow}}

\long\def\ignorethis#1{}

\newcommand{\ourmethod}{GenRC}

\usepackage[accsupp]{axessibility}  

\usepackage[pagebackref,breaklinks,colorlinks,citecolor=eccvblue]{hyperref}

\usepackage{orcidlink}

\begin{document}

\title{\ourmethod: Generative 3D Room Completion from Sparse Image Collections} 

\titlerunning{\ourmethod}

\author{Ming-Feng Li\inst{1} \and
Yueh-Feng Ku\inst{2} \and
Hong-Xuan Yen\inst{2} \and
Chi Liu\inst{4} \and \\
Yu-Lun Liu\inst{3}\and
Albert Y. C. Chen\inst{4}\and
Cheng-Hao Kuo\inst{4}\and
Min Sun\inst{2,4}}

\authorrunning{Li et al.}


\institute{Carnegie Mellon University \and
National Tsing Hua University \and National Yang Ming Chiao Tung University \and Amazon
}

\maketitle

\begin{abstract}

Sparse RGBD scene completion is a challenging task especially when considering consistent textures and geometries throughout the entire scene. Different from existing solutions that rely on human-designed text prompts or predefined camera trajectories, we propose {\ourmethod}, an automated training-free pipeline to complete a room-scale 3D mesh with high-fidelity textures. 
To achieve this, we first project the sparse RGBD images to a highly incomplete 3D mesh. Instead of iteratively generating novel views to fill in the void, we utilized our proposed E-Diffusion to generate a view-consistent panoramic RGBD image which ensures global geometry and appearance consistency. Furthermore, we maintain the input-output scene stylistic consistency through textual inversion to replace human-designed text prompts. To bridge the domain gap among datasets, E-Diffusion leverages models trained on large-scale datasets to generate diverse appearances. {\ourmethod} outperforms state-of-the-art methods under most appearance and geometric metrics on ScanNet and ARKitScenes datasets, even though {\ourmethod} is not trained on these datasets nor using predefined camera trajectories. \\ Project page: \href{https://minfenli.github.io/GenRC}{this https URL}.


\end{abstract}
\section{Introduction}
\label{sec:intro}

3D Scenes are essential in a diverse range of applications, including virtual reality, augmented reality, computer graphics, and game development. Conventional approaches to acquire a 3D scene are formulated as reconstruction by fitting multiple observations such as point clouds or multi-view images. Starting from the seminal works~\cite{Park_2019_CVPR, NeRF}, neural implicit representations have become the most popular type of methods as they demonstrated great accuracy and flexibility in reconstruction and rendering. However, these approaches typically require dense observation of the scene for high-quality interpolation but struggle to extrapolate (or generate) the missing part of the scene. 

\begin{figure}[t]
\centering
\includegraphics[width=0.98\columnwidth]{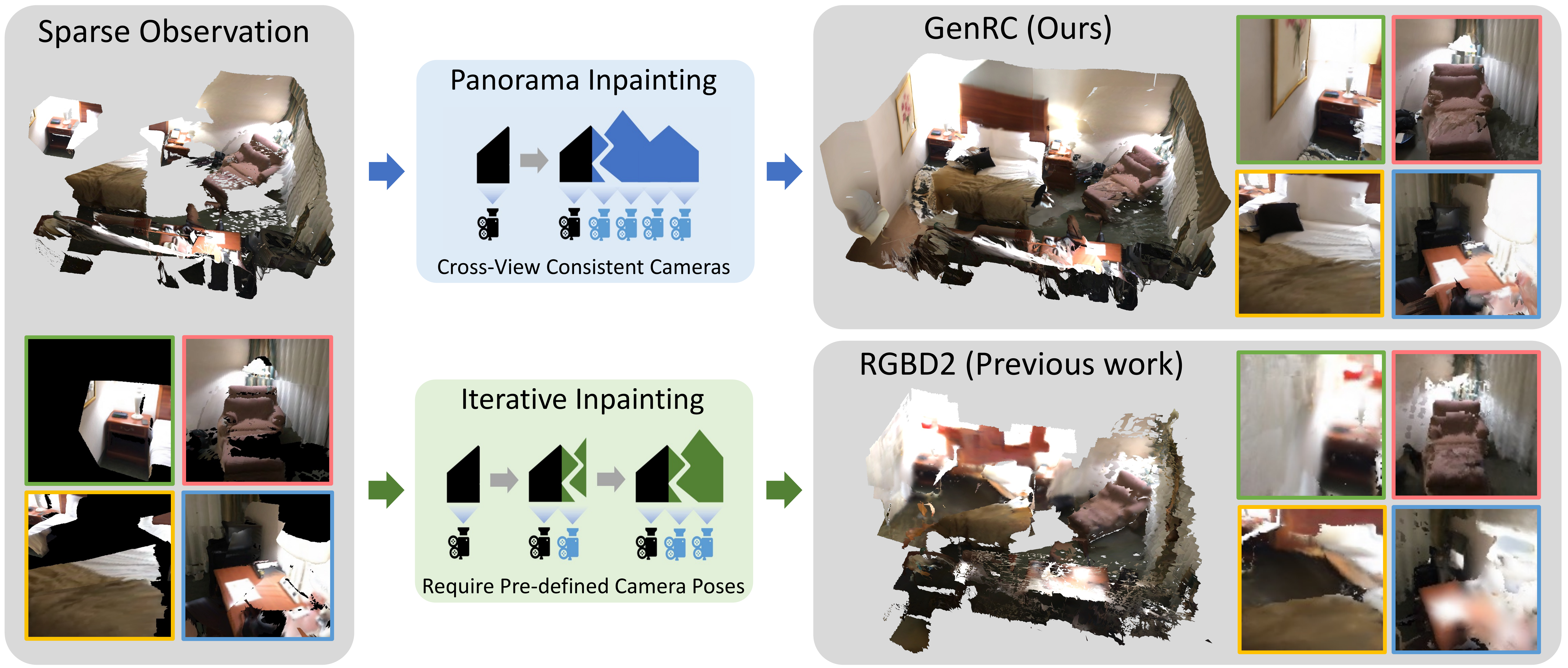}
\caption{
    \textbf{Scene-level 3D mesh generation.}
    {\ourmethod} (the blue path) directly generates a cross-view consistent panorama to complete the main portion of a scene, unlike the iterative methods (the green path) demonstrated in~\cite{lei2023rgbd2,hollein2023text2room} which require designed camera trajectories. {\ourmethod} can produce a comprehensive room-scale mesh with high-fidelity texture, even when provided with sparse RGBD observations. Compared with the previous method RGBD2~\cite{lei2023rgbd2}, {\ourmethod} excels in generating more complete meshes and high-fidelity images.
}
\vspace{-0.8cm}
\label{fig:teaser}
\end{figure}




In this work, we explore a generative method for completing a 3D room with a sparse collection of RGBD images (refer to the sparse observation shown in~\cref{fig:teaser}).
This task is challenging because the generation of absent parts of the scene is in 3D, which requires cross-view consistency in both appearance and geometry. 
To tackle this challenge, RGBD2~\cite{lei2023rgbd2} constructed a diffusion model trained on RGBD data from the ScanNet dataset~\cite{dai2017scannet}. This model was then utilized to inpaint the missing parts within a scene iteratively along the poses of a given camera trajectory. Nevertheless, constrained by the limited quantity and diversity of training images, RGBD2 can only produce scene structures with low fidelity, and their visual styles closely resemble the training data from ScanNet. Additionally, since RGBD2 adopted an iterative approach to synthesize novel-view images with adjacent camera poses, it may produce cross-view inconsistent results very sensitive to the camera trajectory. Hence, RGBD2 requires manually selected camera trajectory to be provided as input.

Inspired by recent works to generate high-fidelity images using 2D text-to-image models~\cite{nichol2021glide, rombach2022high, saharia2022photorealistic, dalle2}, we propose to leverage foundational diffusion models (e.g., Stable Diffusion~\cite{rombach2022high}) for the completion task and design an automated and training-free pipeline to complete posed RGBD images to a room-scale 3D mesh without the need of human-designed text prompts and predefined camera trajectories. Our method comprises four key steps: (1) Firstly, we extract text embeddings as a token to represent the style of provided RGBD images via textual inversion (\cref{sec:textual inversion}), and the token will be utilized in text prompts for the text-to-image model. (2) Next, we project the provided RGBD images to a 3D mesh. 
(3) Following that, we render a panoramic image from a selected room center, which contains missing parts of the scene. Due to the unique equirectangular geometry in panoramic images, standard 2D diffusion inpainting cannot be directly applied. We propose Equirectangular-Diffusion (referred to as E-Diffusion) which explicitly enforces equirectangular projection in the diffusion process.   
Our E-Diffusion guided by textual inversion concurrently denoises these images (\cref{sec:multi-view diffusion}) and determine their depth via monocular depth estimation~\cite{bae2022irondepth} (\cref{sec:depth inpainting}). This step results in a cross-view consistent panoramic RGBD image that completes the main portion of the mesh.
(4) Lastly, we sample novel views from the mesh to fill in holes (\cref{sec:mesh completion}).
With the geometric and stylistic consistency guaranteed by E-Diffusion and textual inversion, our method can effectively generate cross-view consistent room structures without human-designed text prompts and camera trajectories, as shown in~\cref{fig:teaser}. Moreover, benefiting from Stable Diffusion trained on large-scale datasets~\cite{schuhmann2022laion}, our method can generate high-fidelity and diverse room structures without any domain-specific training (e.g., training on the ScanNet dataset). 

We evaluate our method on the ScanNet~\cite{dai2017scannet} and ARKitScenes~\cite{baruch2021arkitscenes} datasets. In contrast to RGBD2 trained on the ScanNet dataset, our approach demonstrates superior performance in both color and geometric metrics without the need for fine-tuning. Additionally, to showcase the cross-domain adaptability of our approach, we apply our method and RGBD2, which was trained on the ScanNet, to the ARKitScenes dataset. Our method shows better cross-domain adaptability and robustness in real-world scenarios.
To summarize, our contributions are:
\begin{itemize}
\item Creating cross-view consistent 3D meshes for indoor scenes with sparse RGBD observations through our E-Diffusion (\cref{sec:multi-view diffusion}) and enhancing the stylistic and geometric coherence of scenes via textual inversion (\cref{sec:textual inversion}) and a novel active sampling method (\cref{sec:active sampling}).
\item Generating high-fidelity and diverse room structures on both ScanNet and ARKitScenes by leveraging Stable Diffusions~\cite{rombach2022high} trained on large datasets~\cite{schuhmann2022laion}.
\item Showcasing a training-free automated pipeline for 3D indoor scene generation without requiring human-designed text prompts or predefined camera trajectories, setting it apart from previous methods.
\end{itemize}

\section{Related Work}
\label{sec:related}


\subsection{2D Diffusion Models}
2D image generation has experienced remarkable progress in text-based control (e.g., GLIDE~\cite{nichol2021glide}, DALL·E 2~\cite{dalle2}, Stable Diffusion~\cite{rombach2022high}, and Imagen~\cite{saharia2022photorealistic}), primarily driven by large-scale text-image datasets~\cite{schuhmann2022laion} and novel diffusion model architectures~\cite{ho2020denoising, rombach2022high}.
In addition to text descriptions, some studies also explored ways to control the output results through other kinds of input conditions such as reference images~\cite{saharia2022palette, liu2023zero, yang2023paint}, sketches~\cite{zhang2023adding, voynov2023sketch}, or incomplete 3D shapes~\cite{cheng2023sdfusion, muller2023diffrf}, etc. To generate precise text descriptions that describe user-desired outputs,~\cite{gal2022image} provides a framework for converting the concepts of images into text embeddings through optimization. 

\begin{figure*}[t]
\centering
\includegraphics[width=\columnwidth]{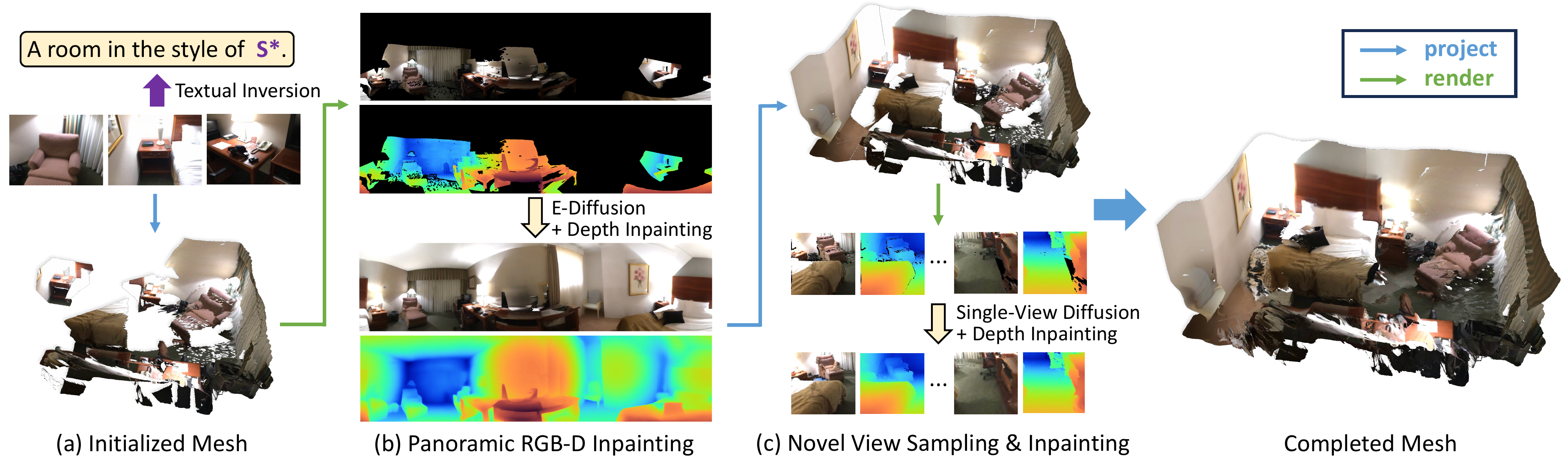}
\caption{
    \textbf{Pipeline of {\ourmethod}:} (a) Firstly, we extract text embeddings as a token to represent the style of provided RGBD images via textual inversion. Next, we project these images to a 3D mesh. (b) Following that, we render a panorama from a plausible room center and use equirectangular projection to render various viewpoints of the scene from the panoramic image. Then, we propose E-Diffusion that satisfies equirectangular geometry to concurrently denoise these images and determine their depth via monocular depth estimation, resulting in a cross-view consistent panoramic RGBD image. (c) Lastly, we sample novel views from the mesh to fill in the remaining holes.
}
\vspace{-0.4cm}
\label{fig:pipeline}
\end{figure*}

\subsection{3D Shape Generation}
Many methods~\cite{gao2022get3d, shue20233d, anciuk2023renderdiff, cheng2023sdfusion, erkocc2023hyperdiffusion} have demonstrated object-level 3D generation on a small number of existing 3D datasets such as ShapeNet~\cite{chang2015shapenet}, showing their capabilities to reconstruct 3D shapes of objects. However, due to the limited scale and diversity of these datasets, these methods can only generate simple shapes and a limited number of classes. To overcome the scarcity of 3D datasets, recent approaches~\cite{poole2022dreamfusion, wang2023score, metzer2023latent, lin2023magic3d, tang2023dreamgaussian, liu2023zero, liu2023one, kasten2023point} expanded powerful 2D text-to-image models, such as Stable Diffusion~\cite{rombach2022high}, to 3D shape generation tasks. These approaches leverage pre-trained image diffusion models as priors, optimizing 3D models such as Neural Radiance Fields (NeRFs) via Score Distillation Sampling (SDS). Nevertheless, these methods encounter difficulties in processing large-scale 3D scenes with fine-grained textures due to the limited capacities of implicit representations. In contrast,~\cite{hollein2023text2room, song2023roomdreamer, fridman2023scenescape, lei2023rgbd2} employ explicit representations, such as meshes, to generate or process room-scale 3D scenes and demonstrate high-fidelity visual details. Nevertheless, most of these methods require additional forms of guidance as input to yield ideal results, including detailed text prompts~\cite{hollein2023text2room, fridman2023scenescape}, carefully designed camera strategies~\cite{hollein2023text2room, lei2023rgbd2}, or providing initial meshes~\cite{song2023roomdreamer}.

\subsection{3D-consistent Scene Synthesis}
Recent studies of neural implicit representations~\cite{NeRF, Park_2019_CVPR} have showcased their ability to produce high-quality reconstructions and synthesize novel views. While these methods often rely on a substantial number of overlapping images for high-quality interpolation, they struggle when it comes to extrapolating missing parts of a scene. In contrast, studies of perpetual view generation~\cite{li2022infinitenature, liu2021infinite, cai2023diffdreamer, fridman2023scenescape} aim to generate unseen parts of a scene, performing the synthesis of videos with a single RGB image as the start. However, these approaches only ensure continuity of appearance but not geometric consistency across different views. 

\subsection{Multi-view Diffusion Models}
To generate images from various viewpoints of a scene while ensuring consistent cross-view appearances, multi-view diffusion model~\cite{bar2023multidiffusion} harnessed diffusion models trained for 2D images and incorporated mechanisms to share local features of neighboring 2D images. In each denoising step, the model denoises the neighboring 2D images independently and interpolates the latents of their overlapping regions. These mechanisms guarantee the appearance consistency of generated 2D images across their overlapping areas. 
However, \cite{bar2023multidiffusion} cannot handle the special equirectangular geometry to produce geometrically correct panoramic images. Recently, \cite{tang2023mvdiffusion, wu2023ipoldm} proposed to train diffusion models on panoramic datasets (Structured3D~\cite{zheng2020structured3d} and Matterport3D~\cite{Matterport3D}, respectively) for panorama inpainting. Due to the limited volume and diversity in panoramic datasets, these methods have not been evaluated on cross-datasets to highlight their generalizability.
In this paper, we want to leverage powerful pre-trained diffusion models but also have the ability to handle equirectangular geometry.
Our E-Diffusion (see \cref{sec:method}) enforces the equirectangular geometry at the beginning steps of the multi-view diffusion process to get the scene geometry correct; then, we apply Texture Refinement diffusion process to enhance the local image quality.



\section{Method}
\label{sec:method}

{\ourmethod} generates a complete 3D mesh with high-fidelity texture, conditioning on sparse RGBD observations. Specifically, given $N$ RGB images $\{\mathbf{I}_i\}_{i=1}^{N}$, their depth maps $\{\mathbf{D}_i\}_{i=1}^{N}$ and associated camera poses $\{\mathbf{P}_i\}_{i=1}^{N}$, our method can generate a complete 3D mesh $M=(V, C, S)$ with the vertices $V$, vertex colors $C$, and the faces $S$. 
The core idea of our approach is to initially generate a cross-view consistent panoramic RGBD image that completes a main portion of the scene and then generate separate novel views to fill the remaining holes within the room. We describe our pipeline and components below.

\subsection{Pipeline Overview}
Our pipeline as show in~\cref{fig:pipeline} consists of the following steps:
(1) In~\cref{fig:pipeline}(a)-Top, we extract text embedding as a token to represent the style of provided RGBD images via textual inversion (see~\cref{sec:textual inversion}). The token will be utilized in text prompts as input to guide the inpainting. At the same time, we initialize a mesh with the given RGBD observations and camera poses by projecting RGB values to 3D and connecting neighboring pixels as triangles (see~\cref{fig:pipeline}(a)-Bottom). (2) Next, we render a panoramic image from a selected room center, which contains missing parts of the scene (black parts in~\cref{fig:pipeline}(b)-Top). (3) Then, in~\cref{fig:pipeline}(b)-Bottom), we applied our proposed E-Diffusion (\cref{sec:multi-view diffusion}) to inpaint the missing parts in the panoramic image, and we determine their depth via monocular depth estimation~\cite{bae2022irondepth} (\cref{sec:depth inpainting}). This step results in a cross-view consistent panoramic RGBD image that completes the main portion of the mesh.
As the most critical step to ensure final completion quality, we leverage the sampling capability of E-Diffusion and introduce an active sampling method (Sec.~\ref{sec:active sampling}) to pick an RGBD panorama that best matches the given geometry from multiple samples. (4) Lastly, we sample novel views from the mesh to fill in the remaining disconnected holes in an iterative manner (\cref{sec:mesh completion}).

\subsection{Preliminary: Inpainting 3D Meshes}
\label{sec:inpainting diffusion}
We first define the task of inpainting 2D images.
Given an image $\mathbf{I}$, an inpainting mask $\mathbf{m}$ and a text prompt $\mathbf{T}$ as input, the text-to-image model $G$ can generate a completed image $\mathbf{I}'$ by filling in the areas specified by $\mathbf{m}$  with the appropriate appearance. The process is represented as $\mathbf{I}'= G(\mathbf{I}, \mathbf{m}, \mathbf{T})$.
Inpainting a 3D mesh can be decomposed into inpainting many 2D images assuming each 2D image captures a portion of the missing surfaces on the 3D mesh.
However, the key challenge is that these 2D images are typically interconnected since a missing surface is often not completely observed by a single 2D image.
Recent methods~\cite{hollein2023text2room, lei2023rgbd2} 
address this challenge by iteratively inpainting 2D images along a predefined camera trajectory comprising $L$ poses $\{\mathbf{\hat{P}}_i\}_{i=1}^{L}$. At each step, they render one image $\mathbf{\hat{I}}_i$ from a novel view $\mathbf{\hat{P}}_i$ and generate a mask $\mathbf{m}_i$ indicating which pixels are covered by the mesh. Then, they utilize 2D diffusion models to get the inpainted image $\mathbf{\hat{I}'}_i$ and determine its depth map $\mathbf{\hat{D}}_t'$ with monocular depth estimation models.
However, the completion quality of the iterative methods depends on the camera trajectory significantly.
Hence, we propose a straightforward yet efficient approach to inpaint a panoramic image that completes a significant portion of the mesh with cross-view consistency.

\subsection{Preliminary: Textual Inversion}
\label{sec:textual inversion}

Using text descriptions to precisely control the style or detailed content of a generated image can be challenging.
This task can be significantly simplified once a reference image is given. \cite{gal2022image} introduced \emph{textual inversion} to convert a reference image into a token embedding as a textual token that represents the concepts of the image. Given a set of images from the same scene, we utilize \emph{textual inversion} to extract the token $S^*$ that best describes the style of these images. 
We extract the token $S^*$ for each scene from $N$ given RGB images $\{\mathbf{I}_i\}_{i=1}^{N}$ and then utilize the extracted $S^*$ to describe the style of a room in text prompts. Thanks to textual inversion for stylistic coherency, our method could constantly use a fixed input prompt, ``a simple and clean room in the style of $S^*$.'', for our automated pipeline and generate images that have the closest semantics with input images without requiring human-designed text prompts.

\begin{figure}[t]
\centering
\begin{subfigure}{0.35\linewidth}
    \includegraphics[width=1.02\columnwidth]
        {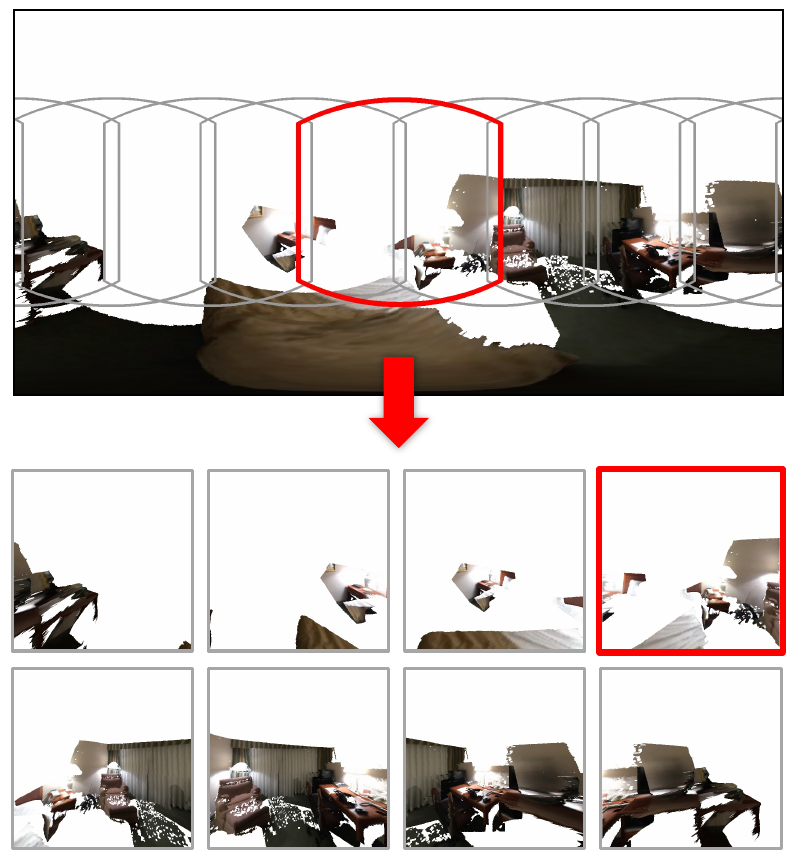}
    \caption{Panorama to perspective images via equirectangular projection}
    \label{fig:multiview_diffusion-a} 
\end{subfigure}
\hspace{0.15cm}
\begin{subfigure}{0.5\linewidth}
    \includegraphics[width=\columnwidth]
        {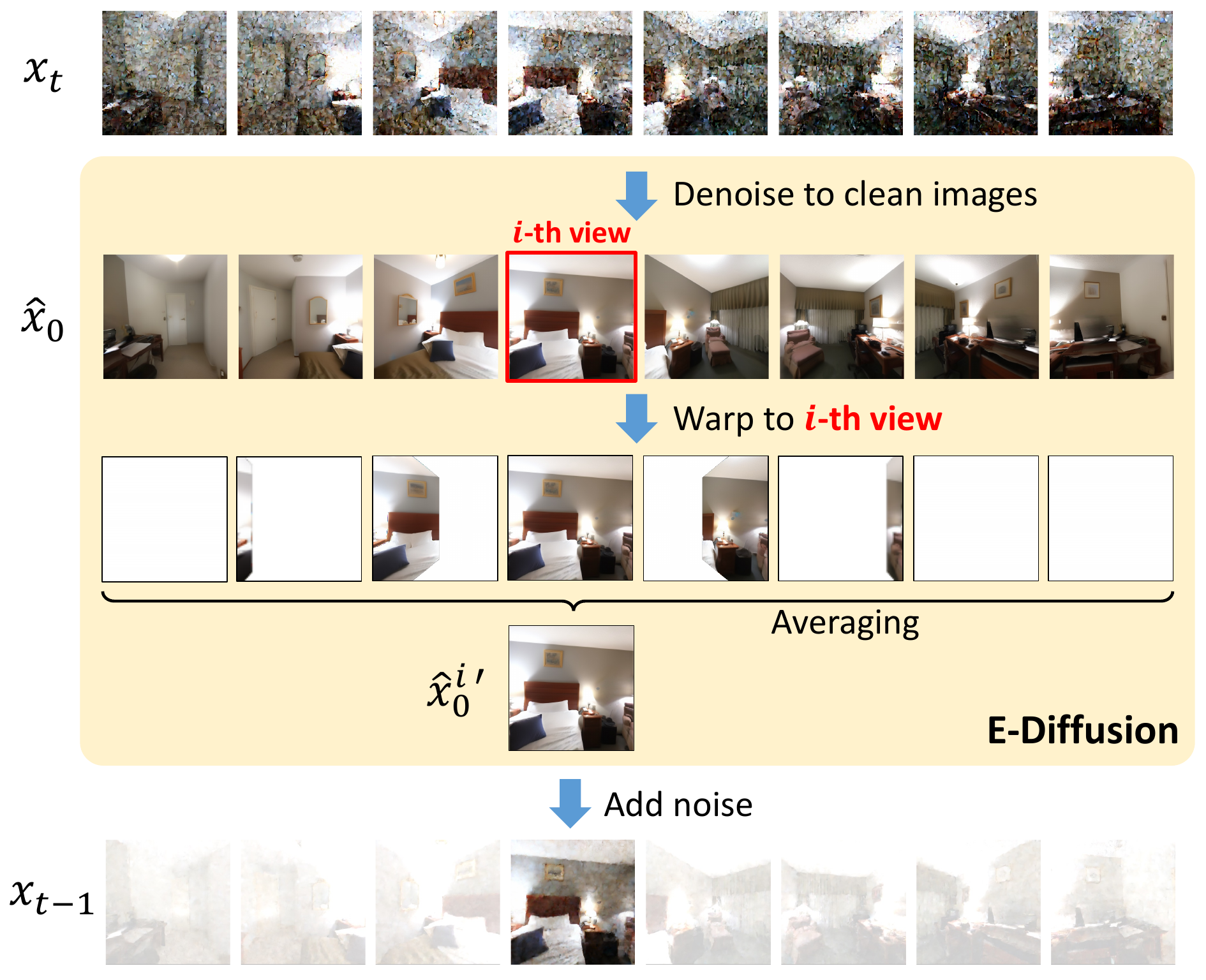}
    \caption{One denoising step of E-Diffusion}
    \label{fig:multiview_diffusion-b}
\end{subfigure}
\vspace{-0.2cm}
\caption{
    \textbf{Multi-view diffusion with equirectangular geometry.} (a) Given an incomplete panoramic image, we first obtain several incomplete perspective images via equirectangular projection. (b) To denoise a perspective image at $i$-th view for one step, we first denoise all images to clean images and warp all the images to $i$-th view to get an averaged image. Then, we add random noise back to the averaged image to get a perspective image which is denoised for one step. Note that while we use images in RGB space here for illustration, the entire process is operated in latent space.
}
\vspace{-0.55cm}
\label{fig:multiview_diffusion}
\end{figure}

\subsection{Panorama Inpainting with Equirectangular-Diffusion}
\label{sec:multi-view diffusion}

\begin{figure*}[t]
\centering
\begin{subfigure}{0.32\linewidth}
    \includegraphics[width=\columnwidth]{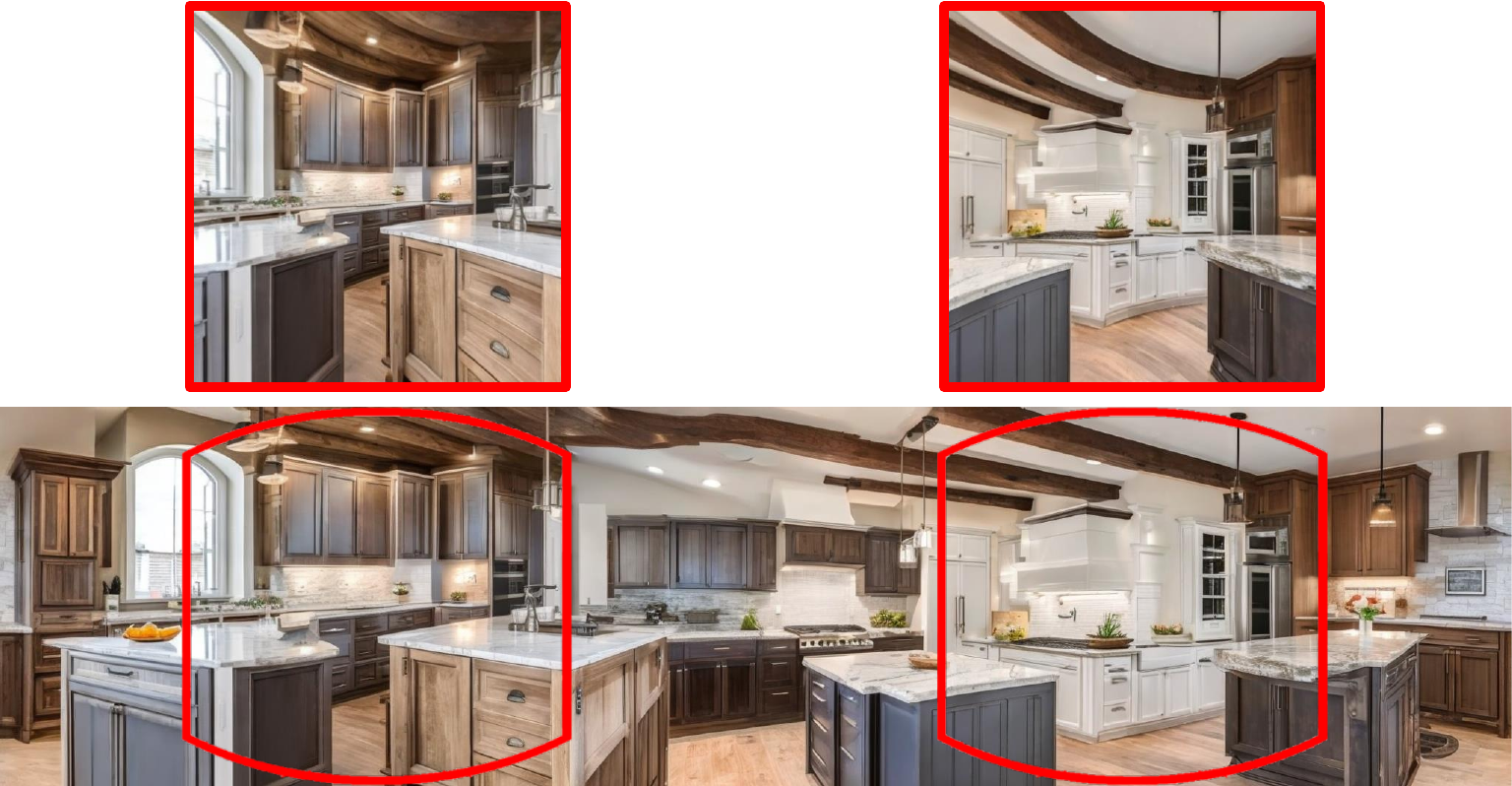}    \caption{MultiDiffusion~\cite{bar2023multidiffusion}}
    \label{fig:pano_comparison-a}
\end{subfigure}
\begin{subfigure}{0.32\linewidth}
    \includegraphics[width=\columnwidth]{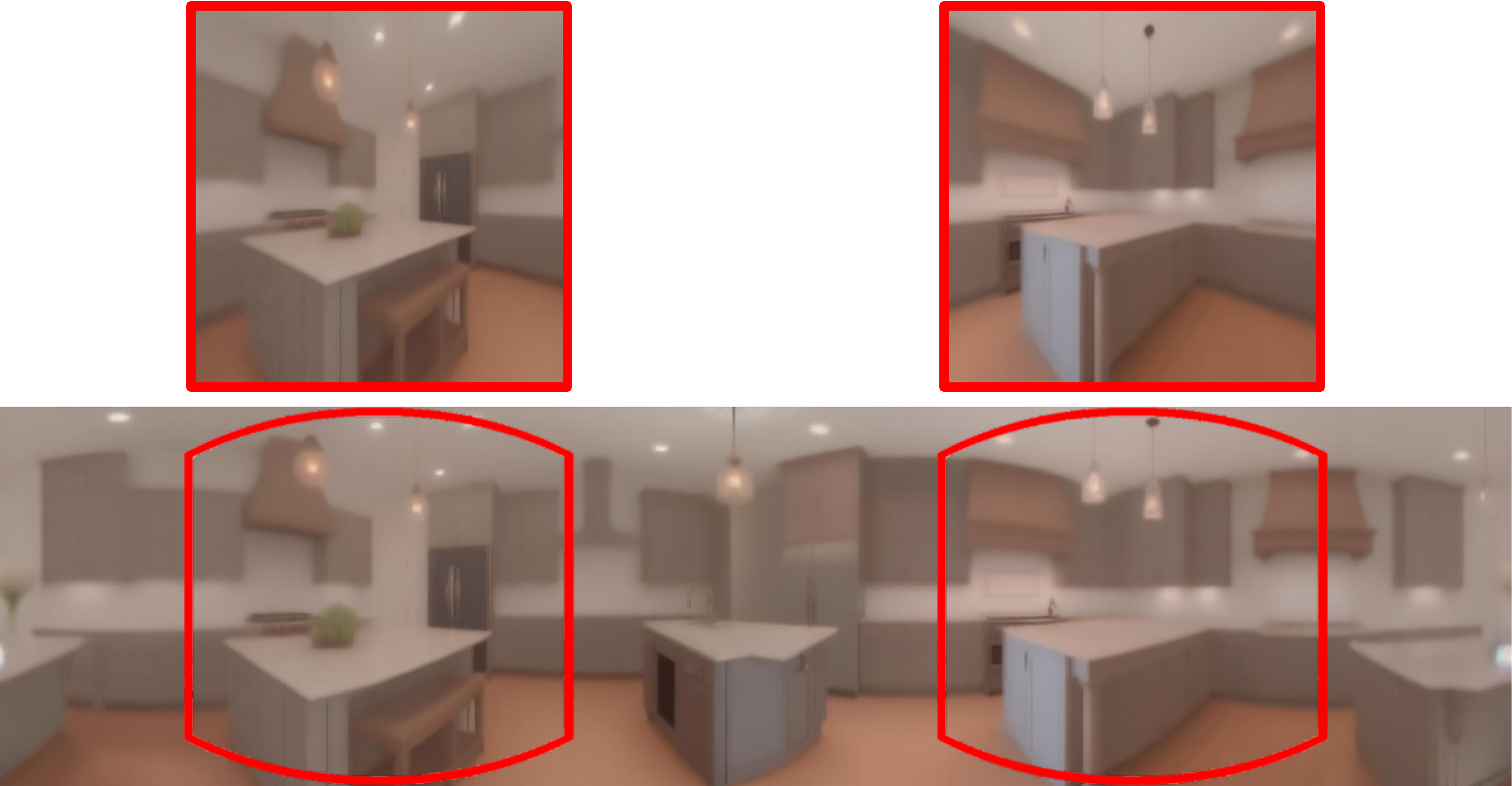}
    \caption{E-Diffusion w/o TR}
    \label{fig:pano_comparison-b}
\end{subfigure}
\begin{subfigure}{0.32\linewidth}
    \includegraphics[width=\columnwidth]{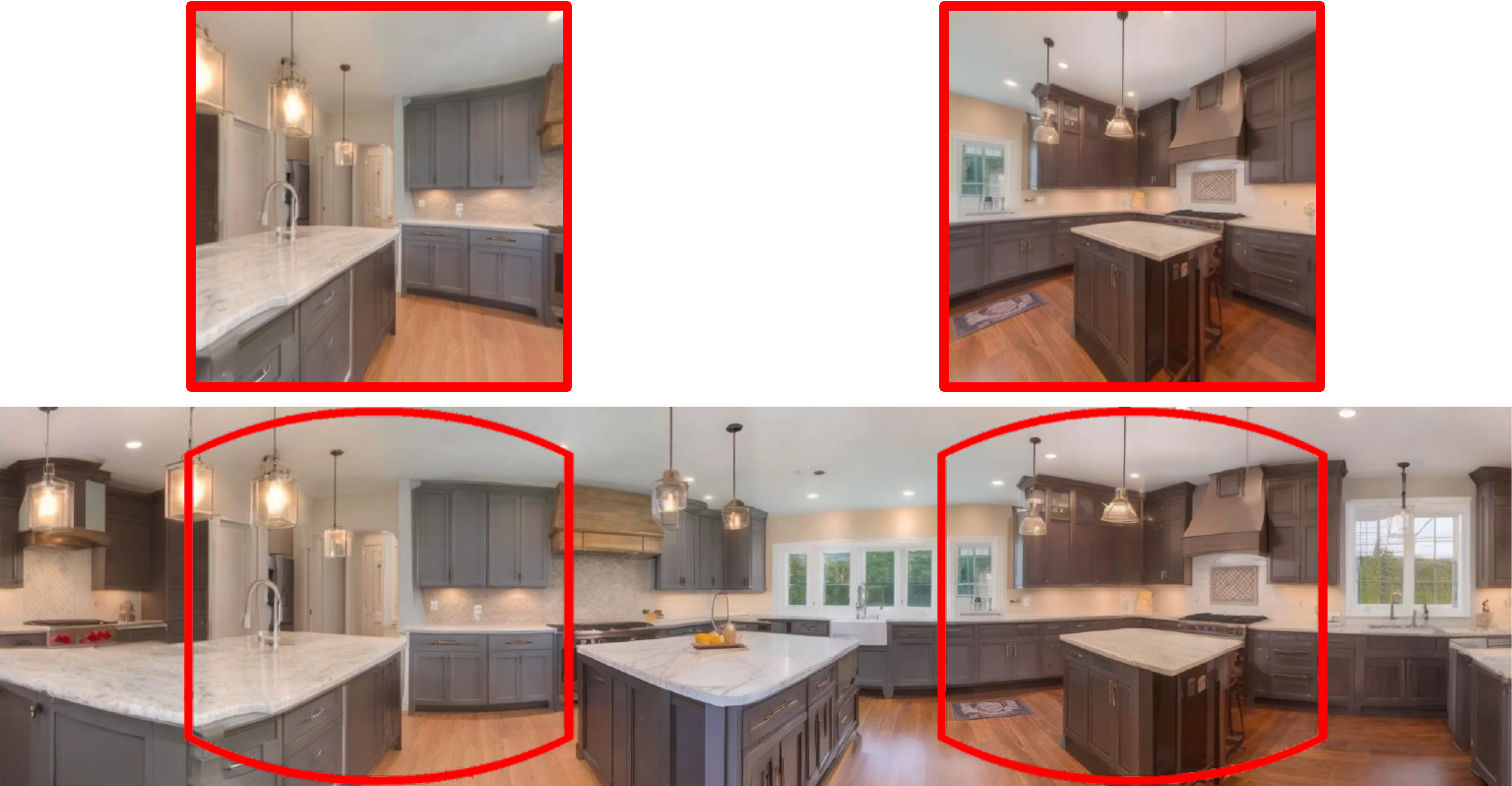}
    \caption{E-Diffusion}
    \label{fig:pano_comparison-c}
\end{subfigure}
\caption{
    \textbf{Comparision of methods for panorama generation.} We crop two regions on each panorama and project them to perspective views (the red blocks above). (a) MultiDiffusion~\cite{bar2023multidiffusion} can produce a high-resolution image. However, it doesn't satisfy the geometry of equirectangular projection (e.g., the straight lines on the ceiling in the panorama transforming into unrealistic curves in the perspective view). (b) Our proposed E-Diffusion (\cref{sec:multi-view diffusion}) can generate a panorama that preserves the equirectangular geometry. But without Texture Refinement (TR), the result looks blurry. (c) Applying the last 20 denoising steps for Texture Refinement (TR), our approach achieves the generation of a high-fidelity and high-resolution panorama that adheres to equirectangular geometry.
}
\vspace{-0.4cm}
\label{fig:pano_comparison}
\end{figure*}

A panoramic image is not simply equal to a perspective image with higher resolution. For instance, in a panorama, horizontal lines located in the top and bottom regions should appear curved, not straight, as they should be stretched relative to those in the middle region (see~\cref{fig:pano_comparison-a}). Hence, pre-trained diffusion models for perspective images are not suitable for panorama inpainting. However, a panorama can be decomposed into multiple perspective images (see \cref{fig:multiview_diffusion-a}) since a panorama uses equirectangular projection to represent a spherical surface, and the spherical surface can be represented by multiple 2D perspective images.
Hence, the key is to inpaint multiple perspective images while preserving cross-view consistency under equirectangular projection.
We propose Equirectangular-Diffusion (referred to as E-Diffusion), a modified MultiDiffusion~\cite{bar2023multidiffusion} for panorama inpainting that considers equirectangular geometry.

To inpaint a panorama with equirectangular geometry, E-Diffusion considers a set of overlapping views represented as $M$ latent images $\{x^i\}_{i=1}^{M}$ with the resolution of 64 $\times$ 64 pixels, whose camera poses share the same position but have different orientations. For each $i$-th view, $\{x^i_{T},...,x^i_{0}\}_{i=1}^{M}$ is then defined as a series of noisy latent images produced by $T$ steps of the reverse diffusion process (see \cref{fig:multiview_diffusion-b}). $\{x^i_{T}\}_{i=1}^{M}$ are initialized as Gaussian noise and $\{x^i_{0}\}_{i=1}^{M}$ are the finally inpainted images. During each step $t$ of the reverse diffusion process, we first predict noise $\epsilon_{\theta}$ of $x^i_{t}$ and obtain $\hat{x}^i_{0}$ for each $i$-th view:
\begin{gather}\label{eq_x0}
    \hat{x}^i_{0} = \dfrac{x^i_{t} - \sqrt{1-\alpha_{t}}  \epsilon_{\theta}(x^i_{t}, t, \mathbf{I}, \mathbf{m}, \mathbf{T})}{\sqrt{\alpha_{t}}}
\end{gather}
where $\mathbf{I}$ and $\mathbf{m}$ are the reference image and the inpainting mask in the pixel space, and $\mathbf{T}$ serves as the text prompt.

Then, to ensure multi-view consistency, for each $i$-th view of $\hat{x}^i_{0}$, all latent images are warped to this $i$-th view and perform averaging to get $\hat{x}^i_{0}{'}$:
\begin{gather}\label{eq_xt-1}
    \hat{x}^i_{0}{'} = \dfrac{\sum_{j} W_{j\rightarrow i}(x^j_{0})}{\sum_{j} m_{j\rightarrow i}}
\end{gather}
where $W_{j\rightarrow i}$ is the warp operation from $j$-th view to $i$-th view and $m_{j\rightarrow i}$ is a binary mask indicating which pixels are visible after warping.
Finally, $x^i_{t-1}$ is obtained by adding random noise $\epsilon$ to $\hat{x}^i_{0}{'}$:
\begin{gather}\label{eq_xt-1}
    x^i_{t-1} = \sqrt{\alpha_{t-1}} \hat{x}^i_{0}{'} + \sqrt{1-\alpha_{t-1}} \epsilon
\end{gather}


\noindent\textbf{Texture Refinement (TR).} 
Although our proposed method can generate a reasonable panorama with equirectangular geometry, it would be blurry since the high-frequency information is lost after many interpolations during the warping of latent pixels, as shown in~\cref{fig:pano_comparison-b}. 
To address this issue, in the last $F$ denoising steps of the total $T$ steps, we remove the equirectangular projection and use MultiDiffusion to refine more high-frequency texture, as shown in~\cref{fig:pano_comparison-c}.
We use $F=20$ for texture refinement and $T=50$ as the total steps of the reverse diffusion process in this work.

\subsection{Panorama Depth Inpainting}
\label{sec:depth inpainting}
Our panorama depth inpainting method is based on a monocular depth estimation model~\cite{bae2022irondepth}, which can predict an initial depth map for a perspective image and refine the depth map recurrently based on the partial ground-truth depth. 
To ensure the predicted depth has the same scale as the depth rendered from the mesh, 
we (1) align the predicted depth to the rendered depth by finding an optimal scale parameter, as~\cite{hollein2023text2room}, after initial depth prediction and (2) use the rendered depth as ground-truth to refine the predicted depth. 

Similar to how we ensure multi-view consistency in \cref{sec:multi-view diffusion}, for each view, we warp the distance maps from the other perspective views to this view and perform averaging to maintain geometric consistency. The \emph{warp-and-average} operation will be performed after predicting initial depth maps for all perspective images and every time we refine the predicted depth with the rendered depth.
We consider depth maps as the depth from the image plane to the object surface, while distance maps represent the distance from a camera origin to object surfaces. Distance maps can be converted from or back to perspective depth maps with camera intrinsic. Note that within our automated pipeline, the depth estimation model can be substituted with other depth estimation methods that incorporate the inpainting function.

\subsection{Active Sampling}
\label{sec:active sampling}



\begin{figure}[t]
\centering
\begin{subfigure}{0.27\linewidth}
    \includegraphics[width=\columnwidth]{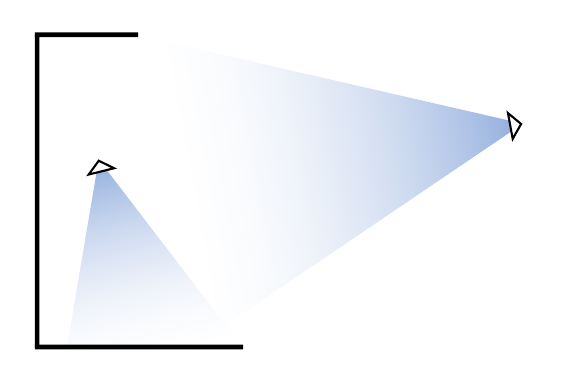}
    \caption{Initialized mesh}
    \label{fig:active-sampling-a}
\end{subfigure}
\begin{subfigure}{0.27\linewidth}
    \includegraphics[width=\columnwidth]{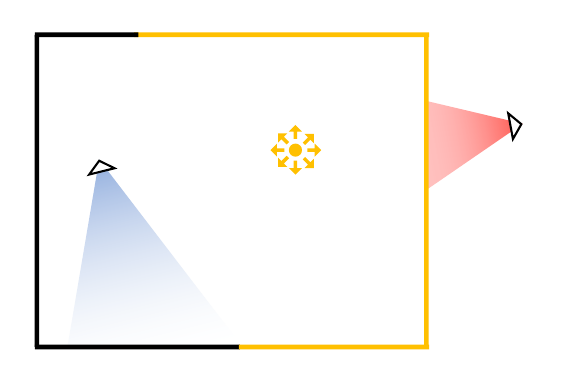}
    \caption{Bad panorama}
    \label{fig:active-sampling-b}
\end{subfigure}
\begin{subfigure}{0.281\linewidth}
    \includegraphics[width=\columnwidth]{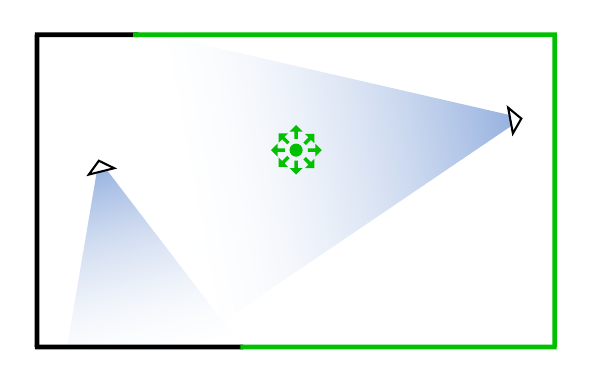}
    \caption{Good panorama} 
    \label{fig:active-sampling-c}
\end{subfigure}
\caption{
    \textbf{Active Sampling.} Given an initialized mesh from two input views as shown in (a), we try to complete the mesh by inpainting the rendered panorama at the room center (yellow and green dots in (b) and (c)). However, input camera views are sometimes blocked by the mesh inpainted from an unreasonable panorama, as shown in (b). To address this issue, our active sampling strategy samples multiple panoramas as candidates and calculates their mean square errors of depth (\cref{eq:mse-active-sampling}) with all input depth maps to pick the best panorama. This strategy prevents us from using bad panoramas that occlude the given camera views.
}
\vspace{-0.4cm}
\label{fig:active-sampling}
\end{figure}

While the proposed method can generate plausible panoramic RGBD images that match the initialized mesh, the input camera views are sometimes blocked by meshes inpainted from bad panoramic images, as shown in~\cref{fig:active-sampling-b}. This is because the input camera poses and the depth maps suggest that the region between input camera positions and projected surfaces must be free space, but this hint is not considered during panorama generation. Following this idea, we introduce a novel active sampling strategy to pick a panoramic RGBD image that best matches the given priors, as shown in~\cref{fig:active-sampling-c}, from multiple panorama samples. Specifically, given a mesh completed by an RGBD panorama, we can compute the mean square error between the $N$ rendered depth maps $\{\mathbf{D}_i^{render}\}_{i=1}^{N}$ and the $N$ input depth maps $\{\mathbf{D}_i\}_{i=1}^{N}$ along input camera poses $\{\mathbf{P}_i\}_{i=1}^{N}$:
\begin{gather}\label{eq:mse-active-sampling}
    MSE_{depth} = \frac{\sum_{i=1}^N (\mathbf{D}_i - \mathbf{D}^{render}_i)^2}{N}
\end{gather} 
If the mean square error is not close to zero, it implies that the sampled panoramic image may have occluded part of the input views. We can sample a set of $A$ panoramic image $\{\mathbf{\bar{I}}_i\}_{i=1}^{A}$ corresponding to their predicted depth maps $\{\mathbf{\bar{D}}_i\}_{i=1}^{A}$ and then pick the one with the minimum mean square error. We set $A=3$ in this work.

\subsection{Mesh Completion}
\label{sec:mesh completion}
After color and depth inpainting, we achieve a temporal mesh that is almost complete by projecting the RGBD panorama back to 3D.
Depending on the use cases, the temporal mesh can be further completed in two different ways. 
To compare with RGBD2 on the benchmark with fixed camera trajectory, we iteratively inpaint the remaining holes along the fixed camera trajectory.
Regardless of the benchmark comparison, we complete the scene by sampling additional camera poses facing existing holes of the mesh. To select optimal camera poses that cover the big holes within the scene while ensuring an adequate portion of the scene for inpainting, we randomly sample multiple camera poses within the scene and select the poses that capture the viewpoint with the highest product of the number of unobserved pixels and the minimum depth, where the number of unobserved pixels infers the region where we want to inpaint and the minimum depth prevents us from selecting a close-up view on small holes. Then, we inpaint the scene with these selected poses iteratively. 
Please see~\cref{sec:supple_meshcomple} for more mesh completion details and visual results.

\section{Experiment}
\label{sec:experiment}

Our approach is evaluated on two indoor datasets: (1) ScanNet~\cite{dai2017scannet} to compare under RGBD2's setting, and (2) ARKitScenes~\cite{baruch2021arkitscenes} to showcase the difference in cross-domain adaptability among {\ourmethod} and other state-of-the-art methods.


\subsection{Setup}
\label{subsec:setup}

\noindent\textbf{Baselines.}
RGBD2~\cite{lei2023rgbd2} is the state-of-the-art 3D scene generation method for sparse RGBD inputs on ScanNet. Its key component is a four-channel (i.e., RGBD) diffusion model trained on ScanNet. In addition, to build a baseline powered by Stable Diffusion~\cite{rombach2022high}, we modify Text2Room (T2R)~\cite{hollein2023text2room}, a text-to-mesh method, to enable it for completing the mesh from sparse RGBD inputs. We refer it to as T2R+RGBD. We use textual inversion as the textual input to T2R+RGBD and initialize the partial 3D mesh for T2R+RGBD to iteratively inpaint RGB and depth.

\noindent\textbf{Datasets.} 
ScanNet~\cite{dai2017scannet} is a well-known RGBD video dataset for indoor generation tasks. 
We follow the same experiment setting shown in~\cite{lei2023rgbd2}, which provides 18 scenes from ScanNet with RGB images and depth maps both in the resolution of $240 \times 180$. At each scene, a certain percentage of images will be uniformly sampled as the sparse set of input images, and the rest of them will be used as testing images for computing metrics. Besides, the in-order camera poses of testing images will be used as fixed camera trajectories for two baseline methods and our mesh completion step in the following experiments. Please refer to~\cref{sup:sensi} for the Sensitivity Analysis of Camera Trajectories which show the impact on each method when in-order camera trajectories are not given.
ARKitScenes~\cite{baruch2021arkitscenes} is a large real-world RGBD video dataset captured using handheld 2020 Apple iPad Pros, making it particularly well-suited for representing real-world use cases. Furthermore, it's worth noting that the depth maps in ARKitScenes exhibit relatively lower density when compared to the ScanNet dataset.
Using the same experimental setup as with the ScanNet dataset, we sampled 20 scenes from the ARKitScenes dataset and filtered out the redundant frames. By conducting experiments on the ARKitScenes dataset, we demonstrate that our method exhibits superior cross-domain adaptability and robustness in real-world scenarios.

\noindent\textbf{Evaluation Metrics.} 
We evaluate the performance by comparing RGB and depth renderings from the completed meshes with ground-truth RGB and depth in the testing sets.
For visual quality, we compute the Peak Signal-to-Noise Ratio (PSNR), Structural Similarity Index Measure (SSIM), and Learned Perceptual Image Patch Similarity~\cite{lpips} (LPIPS). 
For geometry quality, we computed the Mean Squared Error (MSE) on depth maps.
To evaluate if a model can generate images with a similar style to the input RGB images, we compute CLIP Score~\cite{radford2021learning} (CS) for high-level semantic similarity, which is the cosine similarity between the average of the input image CLIP embeddings and the average of the generated image CLIP embeddings. 
Note that quantitative evaluation is conducted along fixed camera trajectories with ground truth RGBD images. This does not show our strength in completing meshes outside the fixed camera trajectories. Please see~\cref{sup:more results} for more qualitative comparisons on panoramic views.

\subsection{Implementation Details.} 
We represent indoor scenes as meshes and utilize Pytorch3D~\cite{johnson2020accelerating} to implement mesh rasterization and fusion. As our text-to-image model, we utilize Stable Diffusion~\cite{rombach2022high} fine-tuned for image inpainting tasks. The resolution of each output image will be $512\times 512$ pixels. As for monocular depth estimation, we employ the IronDepth model~\cite{bae2022irondepth} to estimate the depth of images generated from Stable Diffusion. To identify a plausible room center for panorama inpainting, we calculate the average of camera positions from input images to serve as the room center. The process of creating a panorama from a sparse set of RGBD observations typically takes approximately 3 minutes, while completing one scene consumes roughly half an hour when running on a single RTX 3090 GPU. To compare our method with RGBD2 on the benchmark with fixed camera trajectory, we inpaint the remaining holes of our generated mesh after panorama inpainting along the fixed camera trajectory for our mesh completion step.
Note that the ground truth RGB images and depth maps are at $240 \times 180$ pixels resolution, whereas Stable Diffusion for T2R+RGBD and {\ourmethod} generates high-resolution images at $512 \times 512$ pixels.
Hence, we employ Gaussian blur on rendered images and depth maps of these two methods with the kernel size of $5$ to reduce high-frequency noise and sharp corners when computing metrics.

\subsection{Results on ScanNet}
\label{sec:exp_scannet}

\begin{figure*}[t]
\centering
\includegraphics[width=\linewidth]{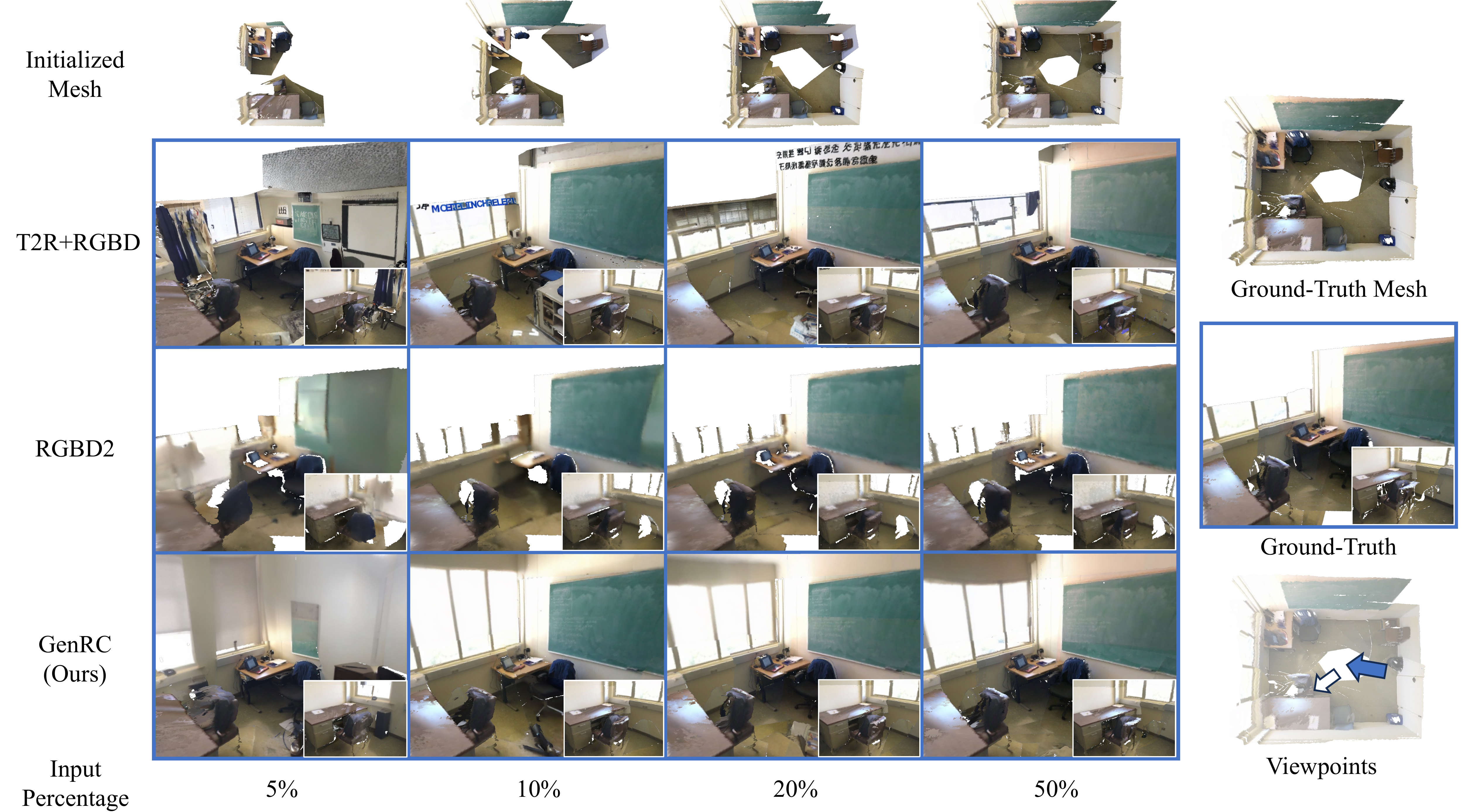}
\vspace{-0.2cm}
\caption{
    \textbf{Comparison with baselines on Scannet.}
    We visualize the generated meshes of each method from two different viewpoints. Leveraging our proposed panorama inpainting technique (\cref{sec:multi-view diffusion}), {\ourmethod} can produce a comprehensive room-scale mesh with high-fidelity texture, even when provided with sparse RGBD observations. In comparison to the prior method RGBD2~\cite{lei2023rgbd2}, {\ourmethod} excels in generating more complete meshes and high-fidelity images. Besides, while T2R+RGBD~\cite{hollein2023text2room} achieves high-fidelity texture, it may generate cross-view inconsistent geometry and artifacts.
}
\vspace{-0.5cm}
\label{fig:scannet}
\end{figure*}

As presented in ~\cref{table:exp_scannet}, {\ourmethod} stands out with higher PSNR and SSIM scores, especially when dealing with sparse RGBD observations. The result demonstrates that {\ourmethod} excels at reconstructing the visual structures of scenes, as shown in~\cref{fig:scannet}. Additionally, in terms of depth mean square error, {\ourmethod} outperforms two baseline methods when sparse observations are provided. This remarkable performance can be attributed to our proposed panorama inpainting technique, as described in~\cref{sec:multi-view diffusion}, which generates cross-view consistent panoramas.
T2R+RGBD shows the lowest performance in geometric metrics, despite achieving competitive scores in feature-level and semantic similarity, such as LPIPS and CS. 
This outcome suggests that T2R+RGBD prioritizes generating high-fidelity images but doesn't effectively address the rational geometry of scenes. Most importantly, as RGBD2 was trained on ScanNet, it performs well when provided with dense RGBD observations as input (i.e., 20\% or 50\%) in PSNR and MSE depth. However, the experimental results show that such performance of RGBD2 does not generalize well across datasets.

\begin{table*}[t]
    \centering
    \caption{\textbf{Quantitative results on ScanNet.} 
    {\ourmethod} stands out with significantly higher PSNR and SSIM scores, especially when dealing with sparse RGBD observations. Additionally, in terms of mean square error in depth estimation, {\ourmethod} outperforms the two baseline methods, particularly when sparse observations are provided.
    }
    \vspace{-0.45cm}
    \resizebox{\linewidth}{!} {
        \begin{tabular}[t]{l*{21}{c}}
             \toprule
               \multirow{3}{*}{\vspace{-0.2cm}Methods} & \multicolumn{12}{c}{Visual} & \multicolumn{4}{c}{Geometric} & \multicolumn{4}{c}{Semantic} \\
              \cmidrule(lr){2-13}\cmidrule(lr){14-17}\cmidrule(lr){18-21}
                 &
                \multicolumn{4}{c}{PSNR$_{\textrm{color}}$(↑)} & 
                \multicolumn{4}{c}{SSIM$_{\textrm{color}}$(↑)} &
                \multicolumn{4}{c}{LPIPS$_{\textrm{color}}$(↓)} &
                \multicolumn{4}{c}{MSE$_{\textrm{depth}}$(↓)} & 
                \multicolumn{4}{c}{CS$_{\textrm{input}}$(↑)} \\
             \cmidrule(lr){2-5}\cmidrule(lr){6-9}\cmidrule(lr){10-13}\cmidrule(lr){14-17}\cmidrule(lr){18-21}
                & 5\% & 10\% & 20\% & 50\%  
                & 5\% & 10\% & 20\% & 50\%  
                & 5\% & 10\% & 20\% & 50\%
                & 5\% & 10\% & 20\% & 50\%  
                & 5\% & 10\% & 20\% & 50\% \\
             \cmidrule(lr){1-21}


                T2R+RGBD~\cite{hollein2023text2room} & 
                12.9 & 15.0 & 16.6 & 17.2 & 
                0.449 & 0.542 & 0.591 & 0.608 & 
                0.573 & 0.492 & 0.444 & 0.423 &
                0.57 & 0.22 & 0.12 & 0.12 &
                0.75 & 0.79 & \textbf{0.81} & \textbf{0.80} \\

                \cmidrule(lr){1-21}
                
                RGBD2~\cite{lei2023rgbd2} & 
                13.7 & 16.0 & 17.5 & \textbf{18.6} & 
                0.501 & 0.562 & 0.598 & 0.609 & 
                0.565 & 0.488 & 0.446 & 0.417 &
                0.38 & 0.15 & \textbf{0.08} & \textbf{0.06} &
                0.69 & 0.71 & 0.72 & 0.73 \\
                
                Ours & 
                \textbf{14.4} & \textbf{16.7} & \textbf{17.6} & 18.2 & 
                \textbf{0.524} & \textbf{0.599} & \textbf{0.628} & \textbf{0.633} & 
                \textbf{0.531} & \textbf{0.441} & \textbf{0.410} & \textbf{0.400} &
                \textbf{0.27} & \textbf{0.13} & 0.09 & 0.09 &
                \textbf{0.77} & \textbf{0.80} & \textbf{0.81} & \textbf{0.80} \\
             \bottomrule
        \end{tabular}
    }
    \label{table:exp_scannet}
\end{table*}

\begin{table*}[t]
    \centering
    \caption{\textbf{Quantitative results on ARKitScenes.} 
    For cross-domain adaptability evaluation, {\ourmethod} demonstrates superior performance in both visual and geometric metrics. Compared with RGBD2 trained on ScanNet, {\ourmethod} consistently outperforms RGBD2 in each metric, which reflects that {\ourmethod} is more suitable for diverse and extensive input data in the real world.
    }
    \vspace{-0.45cm}
    \resizebox{\linewidth}{!} {
        \begin{tabular}[t]{l*{21}{c}}
             \toprule
               \multirow{3}{*}{\vspace{-0.2cm}Methods} & \multicolumn{12}{c}{Visual} & \multicolumn{4}{c}{Geometric} & \multicolumn{4}{c}{Semantic} \\
              \cmidrule(lr){2-13}\cmidrule(lr){14-17}\cmidrule(lr){18-21}
                 &
                \multicolumn{4}{c}{PSNR$_{\textrm{color}}$(↑)} & 
                \multicolumn{4}{c}{SSIM$_{\textrm{color}}$(↑)} &
                \multicolumn{4}{c}{LPIPS$_{\textrm{color}}$(↓)} &
                \multicolumn{4}{c}{MSE$_{\textrm{depth}}$(↓)} & 
                \multicolumn{4}{c}{CS$_{\textrm{input}}$(↑)} \\
             \cmidrule(lr){2-5}\cmidrule(lr){6-9}\cmidrule(lr){10-13}\cmidrule(lr){14-17}\cmidrule(lr){18-21}
                & 5\% & 10\% & 20\% & 50\%  
                & 5\% & 10\% & 20\% & 50\%  
                & 5\% & 10\% & 20\% & 50\%
                & 5\% & 10\% & 20\% & 50\%  
                & 5\% & 10\% & 20\% & 50\% \\
             \cmidrule(lr){1-21}


                T2R+RGBD~\cite{hollein2023text2room} & 
                12.0 & 13.2 & 14.6 & 15.5 & 
                0.408 & 0.458 &  0.521 &  0.551 & 
                0.687 & 0.624 & 0.548 & 0.509 &
                0.66 & 0.69 & 0.35 & 0.27 &
                0.82 & 0.84 & 0.86 & 0.86 \\
                
             \cmidrule(lr){1-21}

                RGBD2~\cite{lei2023rgbd2} & 
                12.2 & 13.9 & 15.2 & 16.6 & 
                0.463 & 0.502 & 0.541 & 0.564 & 
                0.665 & 0.596 & 0.532 & 0.474 &
                0.51 & 0.29 & 0.20 & 0.11 &
                0.78 & 0.80 & 0.81 & 0.81 \\
                
                Ours &
                \textbf{13.2} & \textbf{14.7} & \textbf{15.8} & \textbf{16.8} &
                \textbf{0.504} & \textbf{0.545} & \textbf{0.574} & \textbf{0.592} & 
                \textbf{0.630} & \textbf{0.555} & \textbf{0.499} & \textbf{0.466} &
                \textbf{0.41} & \textbf{0.27} & \textbf{0.14} & \textbf{0.09} &
                \textbf{0.84} & \textbf{0.86} & \textbf{0.87} & \textbf{0.87} \\
             \bottomrule
        \end{tabular}
    }
    \vspace{-0.4cm}
    \label{table:exp_arkit}
\end{table*}

\begin{figure*}[t]
\centering
\includegraphics[width=\linewidth]{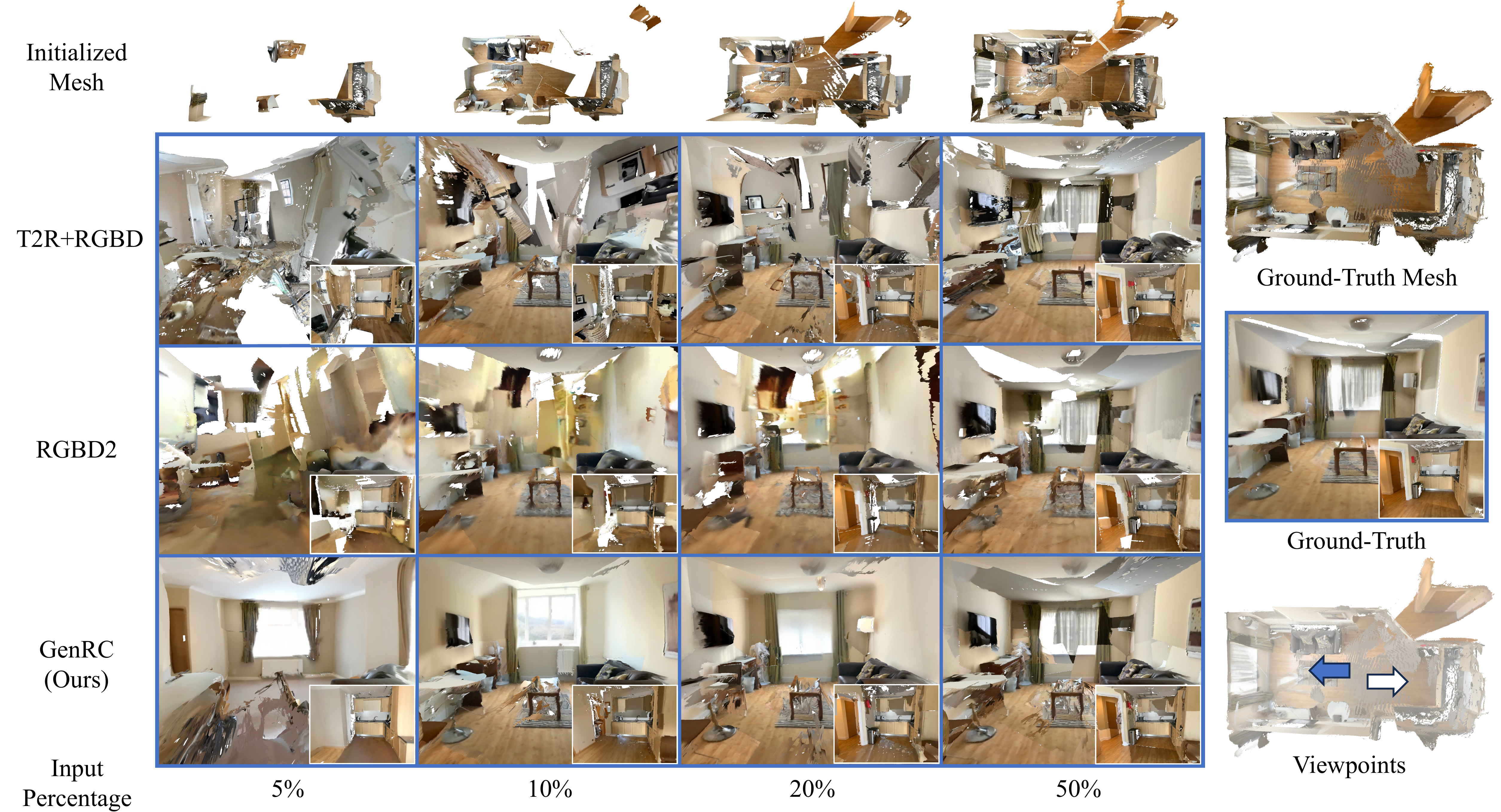}
\vspace{-0.3cm}
\caption{
    \textbf{Comparison with baselines on ArkitScenes.}
    We visualize the generated meshes of each method from two different viewpoints. 
    When the input observations are sparse (5\% and 10\%), both RGBD2 and T2R+RGBD fail to generate reasonable room structures but {\ourmethod} can still produce visually pleasing results.
}
\label{fig:arkit}
\end{figure*}

\subsection{Cross-domain Results on ARKitScenes}
\label{sec:exp_arkits}

We assess the cross-domain adaptability of {\ourmethod} by comparing it with two baseline methods on ARKitScenes~\cite{baruch2021arkitscenes}, without fine-tuning.
As presented in~\cref{table:exp_arkit}, {\ourmethod} demonstrates superior performance in both visual and geometric metrics. In comparison to RGBD2 trained on ScanNet, {\ourmethod} consistently outperforms RGBD2 in all metrics. In addition, while the input observations are sparse (5\% and 10\%), both RGBD2 and T2R+RGBD fail to generate reasonable room structures but {\ourmethod} can still produce visually pleasing meshes, as shown in~\cref{fig:arkit}.
These demonstrate our strength in dealing with sparse observations and show that {\ourmethod} based on Stable Diffusion~\cite{rombach2022high}
trained on large-scale datasets~\cite{schuhmann2022laion} can be more effective for diverse and extensive input data in the real world. 



\begin{table}[t]
    \centering
    \caption{\textbf{Ablation Studies.} 
    Ablation studies reflect the importance of each component in {\ourmethod}. The top three components with the highest scores in each metric are marked in red, orange, and yellow respectively. Please refer to~\cref{sec:ablation} for discussions.
    }
    \vspace{-0.5cm}
    \resizebox{0.6\linewidth}{!} {
        \begin{tabular}[t]{l*{10}{c}}
             \toprule
               \multirow{2}{*}{\vspace{-0.2cm}Methods} &
                \multicolumn{3}{c}{PSNR$_{\textrm{color}}$(↑)} & 
                \multicolumn{3}{c}{MSE$_{\textrm{depth}}$(↓)} &
                \multicolumn{3}{c}{CS$_{\textrm{input}}$(↑)} \\
             \cmidrule(lr){2-4}\cmidrule(lr){5-7}\cmidrule(lr){8-10}
                & 5\% & 10\% & 20\%  
                & 5\% & 10\% & 20\%
                & 5\% & 10\% & 20\%
                \\
             \cmidrule(lr){1-10}

                w/o panorama inpainting & 
                13.7 & 16.1 & 17.4 & 
                0.44 & \orgcell 0.15 & \redcell 0.09 &
                0.781 & 0.805 & 0.807 \\

                w/o E-Diffusion & 
                13.8 & \ylwcell 16.4 & 17.5 & 
                \ylwcell 0.32 & \ylwcell 0.16 & \redcell 0.09 &
                0.765 & 0.799 & 0.808 \\
                
                w/o texture refinement & 
                \orgcell 14.6 & \ylwcell 16.4 & \redcell 17.7 & 
                0.35 & 0.21 & 0.13 &
                0.785 & 0.808 & 0.807\\

                w/o active sampling & 
                \ylwcell 14.5 & \ylwcell 16.4 & \orgcell 17.6 &
                0.46 & 0.19 & 0.14 &            
                \redcell 0.794 & \ylwcell 0.810 & \redcell 0.809 \\

                w/o textual inversion & 
                \redcell 14.7 & \orgcell 16.5 & \orgcell 17.6 & 
                \orgcell 0.30 & \ylwcell 0.16 & 0.12 &
                \ylwcell 0.792 & \orgcell 0.811 & \redcell 0.809 \\


                Ours (full) & 
                14.4 & \redcell 16.7 & \orgcell 17.6 & 
                \redcell 0.27 & \redcell 0.13 & \redcell 0.09 &
                \redcell 0.794 & \redcell 0.812 & \redcell 0.809\\
             \bottomrule
        \end{tabular}
    }
    \label{table:exp_ablation}
    \vspace{-0.5cm}
\end{table}

\subsection{Ablation Studies}
\label{sec:ablation}
We carry out ablation studies to confirm the importance of each component of our method by removing them one at a time in~\cref{table:exp_ablation}. For more visual comparisons, please refer to~\cref{sec:supple_qual}.

\noindent\textbf{Panorama Inpainting.} 
Firstly, we estimate the effectiveness of our proposed panorama inpainting method (\cref{sec:multi-view diffusion}). Without panorama inpainting, the PSNR declines and the depth mean square error increases significantly when the observations are sparse.
The result underlines the importance of utilizing panorama inpainting to complete the main portion of a given mesh at once, instead of iteratively generating novel views to fill in the void.
Next, we compare the ablation studies of two panorama inpainting methods, as described in~\cref{sec:multi-view diffusion}: (1) directly applying MultiDiffusion for panorama inpainting (referred to as w/o E-Diffusion); (2) performing our proposed E-Diffusion without texture refinement (referred to as w/o texture refinement). 
In the case of directly applying MultiDiffusion for panorama inpainting (see~\cref{fig:pano_comparison-a}), the generated panorama doesn't adhere to the equirectangular geometry. Hence, both visual and geometric metrics decrease significantly when the observations are sparse.
In the case without texture refinement (see~\cref{fig:pano_comparison-b}), the blurriness of inpainted panoramas slightly decreases the visual metrics. Also, the mean squared error of depth grows higher because the depth estimation model cannot predict the depth accurately for blurry panoramas without clear corners and details.

\noindent\textbf{Active Sampling.} 
Taking away active sampling yields the highest mean square error of depth, underlining its significance for maintaining geometric consistency with input views. Moreover, active sampling can enhance visual quality by preventing input views from being obstructed or occluded by generated meshes.

\noindent\textbf{Textual Inversion.} 
Textual inversion helps to generate images that show higher semantic similarity with provided images, resulting in higher CLIP scores.

\section{Conclusion}
\label{sec:conclusion}
In this work, we proposed {\ourmethod}, a training-free automated pipeline for 3D indoor scene generation.
By leveraging the powerful pre-trained diffusion model~\cite{rombach2022high} and our proposed E-Diffusion (\cref{sec:multi-view diffusion}) for cross-view consistent panoramas, {\ourmethod} can generate complete room-scale 3D meshes with high-fidelity texture given a sparse collection of RGBD images. Furthermore, we proposed an active sampling manner (\cref{sec:active sampling}) and utilized textual inversion to enhance the geometric and stylistic consistency of scenes.
Notably, {\ourmethod} outperforms state-of-the-art methods under most appearance and geometric metrics on ScanNet and ARKitScenes datasets even though {\ourmethod} was not trained on these datasets.

\section*{Acknowledgements}
This project is supported by the National Science and Technology Council
(NSTC) and Taiwan Computing Cloud (TWCC) under the project NSTC 112-2634-F-002-006 and 113-2221-E-007-104.

%
%
\bibliographystyle{splncs04}
\bibliography{main}

\newpage
\appendix

\section*{Appendix}

\section{Datasets and Metrics}
\label{sup:metric}

\subsection{Data Preprocessing}

\noindent\textbf{ScanNet.}
To prepare data from the ScanNet~\cite{dai2017scannet} dataset, we follow the same experiment setting shown in~\cite{lei2023rgbd2}, which provides 18 scenes from ScanNet with RGB images and depth maps both in the resolution of $240 \times 180$. Specifically, we uniformly sample 18 scenes whose number of views is more than 50 from the testing set following~\cite{lei2023rgbd2}. 

\noindent\textbf{ArkitSences.}
Similarly, we prepare the ArkitSences~\cite{baruch2021arkitscenes} dataset by preprocessing the RGB images and depth maps into the resolution of $240 \times 180$ as the same experiment setting shown in~\cite{lei2023rgbd2}. Then, we uniformly sample 100 views per scene, so the number of percentages in the experiments will be the number of views we used for initializing a mesh (e.g., 5\% means 5 views). For sampling the scenes for evaluation, we consider the official testing set which contains 549 scenes, and select the 189 scenes that were captured without hand-held camera rotation. Finally, we uniformly sample 20 scenes from them for evaluation.

\subsection{Metric Calculation}
We follow~\cite{lei2023rgbd2} to compute evaluation metrics shown in the tables. For each scene, a certain percentage of images (5\%, 10\%, 20\%, or 50\%) will be uniformly sampled for all the images from this scene as input sparse observations, and the rest of them will be used as testing images for computing metrics. Note that the in-order camera poses of testing images will serve as prefined camera trajectories for two baseline methods to inpaint their meshes. Even if there exist holes on the generated meshes of T2R+RGBD and RGBD2, these holes are outside the trajectories which won't influence the quantitative results.

\subsection{Geometric Metrics}
We evaluate the geometric quality by comparing depth renderings from the generated meshes with ground-truth depth maps in the testing sets (see~\cref{subsec:setup}).
In addition, the depth MSE measures the geometric consistency of a scene since ground-truth depth maps come from the same scene and should be geometry-consistent.
In~\cite{lei2023rgbd2}, bi-directional Chamfer Distance is used as an alternative for evaluating the geometric quality. 
However, we observed that the ground-truth mesh generated by back-projecting ground-truth images may not capture the whole scene. As a result, the bi-directional Chamfer Distance metric will also penalize our method when our generated meshes are more complete.
Hence, we report one-directional Chamfer Distance (the distance between each point in the ground-truth mesh and its nearest neighbor in the generated mesh) in~\cref{table:exp_scannet_chamfer_and_completeness}. Our method outperforms other state-of-the-art methods, especially when input observations are sparse.

\begin{table}[t]
    \centering
    \caption{\textbf{Additional quantitative results of geometry quality on ScanNet.} 
    We report one-directional Chamfer Distance (one-directional CD), which shows the distance between each point in the ground-truth mesh and its nearest neighbor in the generated mesh.
    {\ourmethod} can generate more geometrically correct meshes when sparse observations are given (i.e., 5\%, 10\%, 20\%).
    }
    \vspace{-0.5cm}
    \resizebox{0.5\linewidth}{!} {
        \begin{tabular}[t]{l*{5}{c}}
             \toprule
               \multirow{2}{*}{\vspace{-0.2cm}Methods} &
                \multicolumn{4}{c}{One-directional CD(↓)} \\
             \cmidrule(lr){2-5}
                & 5\% & 10\% & 20\% & 50\% \\             
                \cmidrule(lr){1-5}
                T2R+RGBD~\cite{hollein2023text2room}& 
                0.091 & 0.031 & 0.018 & 0.014 \\ 

                \cmidrule(lr){1-5}
                
                RGBD2~\cite{lei2023rgbd2} & 
                0.073 & \textbf{0.018} & 0.011 & \textbf{0.005} \\ 

                Ours & 
                \textbf{0.050} & \textbf{0.018} & \textbf{0.009} & 0.007 \\
             \bottomrule
        \end{tabular}
    }

    \vspace{-0.7cm}
    \label{table:exp_scannet_chamfer_and_completeness}
\end{table}


                


\subsection{Baseline Implementation}
We utilized the official codebases of RGBD2~\cite{lei2023rgbd2} and Text2Room~\cite{ hollein2023text2room} and conducted experiments on them for evaluation. In particular, we modify Text2Room~\cite{hollein2023text2room} as T2R+RGBD to complete the mesh from sparse RGBD inputs by initializing the mesh for Text2Room and utilizing the camera poses of testing images as the predefined camera trajectory. For T2R+RGBD, the text prompt to Stable Diffusion~\cite{rombach2022high} is ``a simple and clean room in the style of $S^*$'', where $S^*$ is the textual token from \emph{textual inversion}.
\section{Method Details}
\label{sec:method_detail}

\subsection{Text Prompt}
We utilize \emph{textual inversion}~\cite{gal2022image} to extract the token $S^*$ to represent the style of a room and use it in text prompts for Stable Diffusion, as described at~\cref{sec:textual inversion}. The following templates are used to extract the token $S^*$ that represents the style of a room: ``a $S^*$ room'', ``the $S^*$ room'', ``one $S^*$ room'', ``a room in the style of $S^*$'', ``the room in the style of $S^*$'', and ``one room in the style of $S^*$''.

Then, we utilize the extracted $S^*$ in text prompts for Stable Diffusion. Note that we use a fixed input prompt:
``a simple and clean room in the style of $S^*$''. for all image inpainting, so {\ourmethod} doesn't require any scene-specific or detailed prompts shown in previous works~\cite{hollein2023text2room, fridman2023scenescape}.

\subsection{E-Diffusion}
For E-Diffusion, we consider 8 rectangular views with the field of view as 98 degrees, which ensures the stitched panorama is fully covered by these views. The noise $\epsilon$ used to obtain $x^i_{t-1}$ in Eq. (3) is sampled randomly from the Gaussian distribution of unit variance every two steps.
For the input of MultiDiffusion~\cite{bar2023multidiffusion} used in texture refinement, we stitch the eight perspective views together as one equirectangular panorama of $2048 \times 1024$ pixels and only keep the region with latitude between -45 and 45 degrees, resulting in a panoramic image of $2048 \times 512$ pixels. For MultiDiffusion, we consider 16 sliding windows with window size of $512 \times 512$ pixels and step size of $128$ pixels.



\subsection{Mesh Completion}
\label{sec:supple_meshcomple}

\begin{figure}[t]
\centering
\begin{subfigure}{0.3\linewidth}
    \includegraphics[width=\columnwidth]{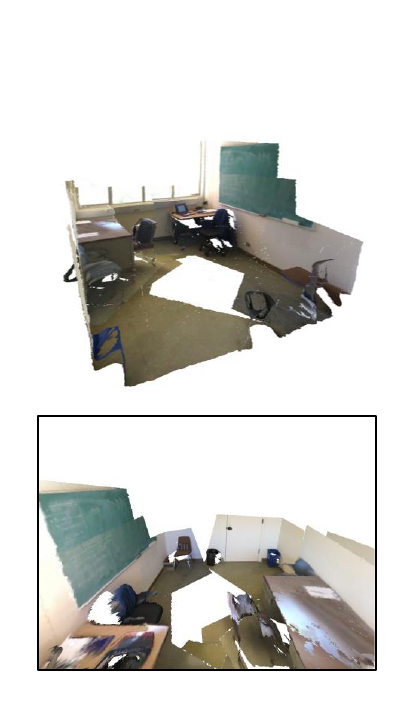}
    \caption{Initialized mesh}
    \label{fig:active-sampling-a}
\end{subfigure}
\begin{subfigure}{0.3\linewidth}
    \includegraphics[width=\columnwidth]{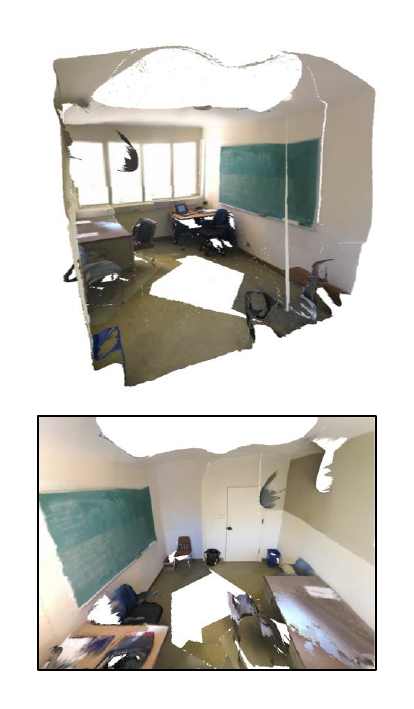}
    \caption{After panorama}
    \label{fig:active-sampling-b}
\end{subfigure}
\begin{subfigure}{0.3\linewidth}
    \includegraphics[width=\columnwidth]{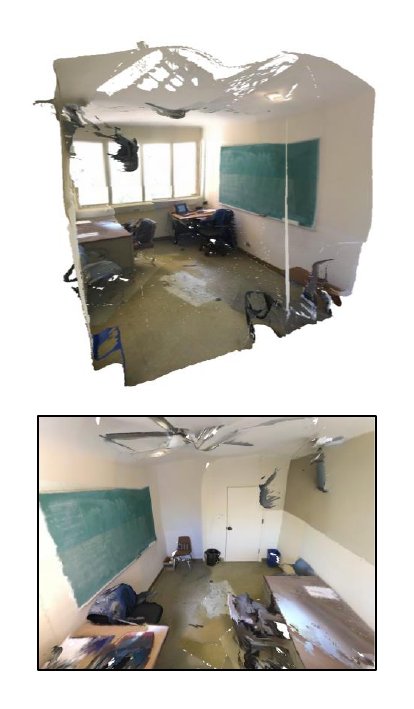}
    \caption{After completion} 
    \label{fig:active-sampling-c}
\end{subfigure}
\vspace{-0.2cm}
\caption{
    \textbf{Mesh completion.}
As described in~\cref{sec:mesh completion}, our mesh completion method can further complete the temporal mesh after RGB and depth panorama inpainting by sampling additional camera poses facing existing holes in the mesh.
}
\vspace{-0.2cm}
\label{fig:completion}
\end{figure}

\begin{table}[t]
    \centering
    \caption{\textbf{Sensitivity Analysis of Camera Trajectories.} 
    The metrics of RGBD2 dramatically decline without in-order camera trajectories which are composed of closely adjacent camera poses. While T2R+RGBD performs a higher depth MSE, given shuffled trajectories, the decreasing visual metric reflects it generates structures that are not similar to the ground truth. However, {\ourmethod} can still effectively generate cross-view consistent room structures even if given camera trajectories are not in order.
    }
    \vspace{-0.4cm}
    \resizebox{0.7\linewidth}{!} {
        \begin{tabular}[t]{l*{8}{c}}
             \toprule
               \multirow{2}{*}{\vspace{-0.2cm}Methods} &
                \multicolumn{2}{c}{PSNR$_{\textrm{color}}$(↑)} & 
                \multicolumn{2}{c}{MSE$_{\textrm{depth}}$(↓)} & 
                \multicolumn{2}{c}{CS$_{\textrm{input}}$(↑)} \\
             \cmidrule(lr){2-3}\cmidrule(lr){4-5}\cmidrule(lr){6-7}
                & in-order & shuffled  
                & in-order & shuffled  
                & in-order & shuffled \\
             \cmidrule(lr){1-7}

                T2R+RGBD~\cite{hollein2023text2room} & 
                11.6 & 11.2 & 0.88 & \textbf{0.76} & 0.72 & \textbf{0.75} \\
                 
             \cmidrule(lr){1-7}
                                 
                RGBD2~\cite{lei2023rgbd2} & 
                12.2 & 10.7 & 0.72 & 1.72 & 0.69 &  0.66 \\

                Ours & 
                \textbf{12.9} & \textbf{12.7} & \textbf{0.59} & 0.77 & \textbf{0.74} & \textbf{0.75} \\

             \bottomrule
        \end{tabular}
    }
    \vspace{-0.4cm}
    \label{table:exp_scannet_shuffle}
\end{table}

We demonstrate our mesh completion method for the generation of a complete room-scale mesh. To this end, we will iteratively select 30 camera poses to patch up the remaining holes in the mesh. To select an optimal camera pose in each iteration, we first find the bounding box that covers the scene mesh and randomly sample 200 camera poses centered within the central 80\% of the bounding box in the horizontal direction and central 10\% of the bounding box in the vertical direction. The elevation angles are between 15 and -15 degrees. Given the inpainting ratio as the ratio of unobserved pixels to total pixels in the rendered image and the back-face ratio as the ratio of pixels that are rendered from the back of the mesh, we filter out the camera poses with inpainting ratio greater than 50\%, back-face ratio greater than $1\%$, or minimum depth less than $1$m. Then, we select the camera poses with the highest product of inpainting ratio and minimum depth. Finally, we move the selected camera poses backward as long as the criteria we use to filter out camera poses are satisfied, which helps include as much information as possible into the field of view. We showcase the effectiveness of our proposed mesh completion method (mentioned in~\cref{sec:mesh completion}) qualitatively in~\cref{fig:completion}.

\begin{figure*}[t]
\centering
\includegraphics[width=\linewidth]{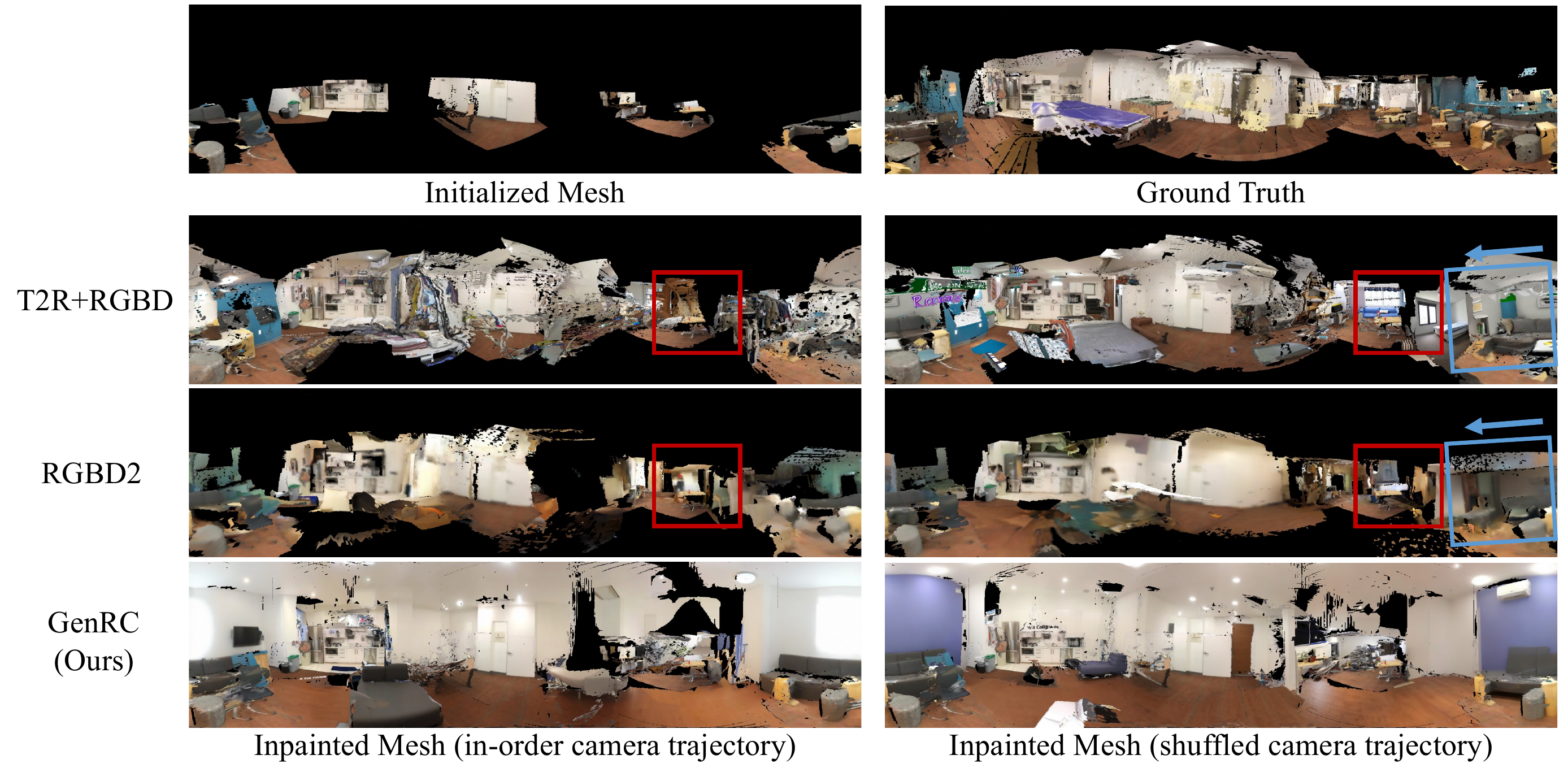}

\vspace{-0.2cm}
\caption{
    \textbf{Sensitivity Analysis of Camera Trajectories.} In comparison to {\ourmethod} which aims to generate a panorama covering most parts of the scene, T2R+RGBD and RGBD2 iteratively generate the scene and require pre-designed camera trajectories composed of closely adjacent camera poses to ensure reasonable cross-view geometry. In addition, these methods should start from a viewpoint where a certain portion of the mesh exists to ensure appearance and geometric consistency. For instance, T2R+RGBD and RGBD2 may produce inconsistent geometries that cannot close the scene (in the red boxes) and produce unreasonable room structures that are not perpendicular to the ground (in the blue boxes).
    In contrast, {\ourmethod} can still generate cross-view consistent and complete room structures even if given camera trajectories are arbitrary.
}
\vspace{-0.5cm}
\label{fig:shuffle}
\end{figure*}

\section{Sensitivity Analysis of Camera Trajectories}
\label{sup:sensi}

In comparison to {\ourmethod} which aims to generate a panorama covering most parts of the scene, RGBD2 and T2R+RGBD iteratively generate the scene following a pre-designed camera trajectory composed of closely adjacent camera poses.
In addition, these methods should start from a viewpoint where a certain portion of the mesh exists to ensure appearance and geometric consistency. 
In this analysis, we test each method on the extremely sparse observations as 3\% on the ScanNet dataset to analyze the sensitivity when the given camera trajectories are not composed of closely adjacent camera poses. To this end, we randomly shuffle the originally ``in-order'' camera trajectories, which are the sequences of camera poses of testing images, to ``shuffled'' camera trajectories. In~\cref{table:exp_scannet_shuffle}, we can observe that, without ``in-order'' camera trajectories composed of closely adjacent camera poses, the visual metrics of both RGBD2 and T2R+RGBD decline while {\ourmethod}'s performance almost remains the same. Especially, the decreasing PSNR and increasing depth mean square error reflect that RGBD2 fails to reconstruct the reasonable appearance and geometry of a scene.
Even if T2R+RGBD shows the higher mean square of depth while using shuffled trajectories, the decreasing visual metric reflects it generates structures that are not similar to the ground truth.
For instance, as shown in~\cref{fig:shuffle}, we can observe that T2R+RGBD and RGBD2 may generate inconsistent geometries that cannot close the scene and produce unreasonable room structures that are not perpendicular to the ground.
Thanks to the continuous geometry provided by panorama inpainting (\cref{sec:multi-view diffusion} and~\cref{sec:depth inpainting}), the big portion of a mesh has been completed after the RGBD inpainting and therefore {\ourmethod} can effectively generate cross-view consistent room structures even if camera trajectories are arbitrary.

\begin{figure*}[t]
\centering
\begin{subfigure}{\linewidth}
    \includegraphics[width=\linewidth]{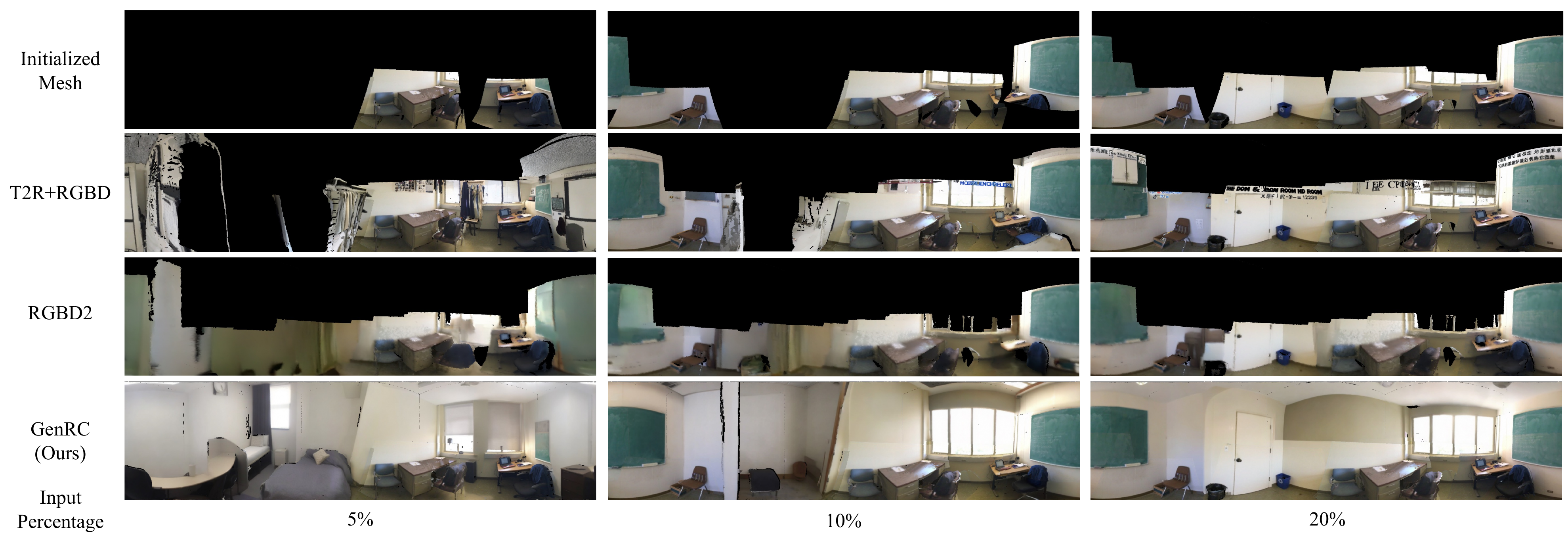}
    \vspace{-0.4cm}
    \caption{Scene 628\_01}
\end{subfigure}

\begin{subfigure}{\linewidth}
    \includegraphics[width=\linewidth]{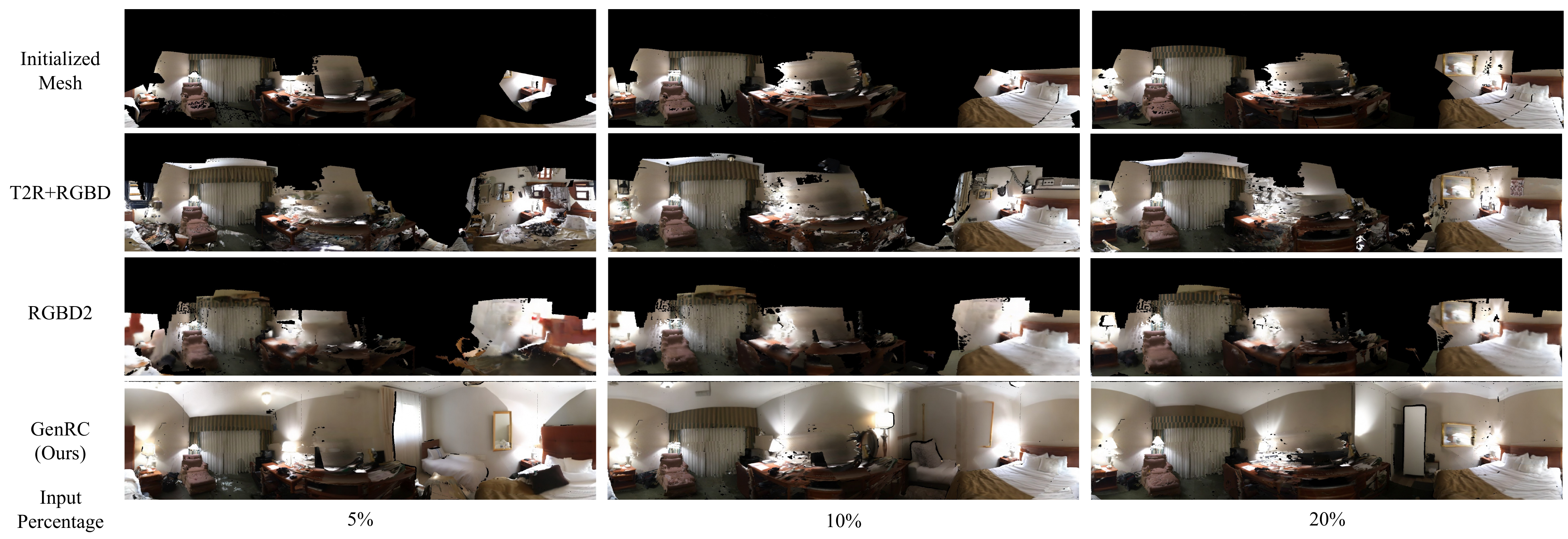}
    \vspace{-0.4cm}
    \caption{Scene 679\_01}
\end{subfigure}
\vspace{-0.6cm}
\caption{
    \textbf{Comparison with baselines on ScanNet.} We project generated meshes to panoramas to demonstrate the portions of meshes that are completed. In comparison to the prior method RGBD2~\cite{lei2023rgbd2}, {\ourmethod} can generate more complete meshes and high-fidelity images due to RGB and depth inpainting of {\ourmethod}. Besides, while T2R+RGBD~\cite{hollein2023text2room} achieves high-fidelity texture, it may generate cross-view inconsistent geometry and artifacts. Please refer to~\cref{sec:exp_scannet} for more quantitative discussions.
}
\label{fig:scannet_panorama}
\vspace{-0.5cm}
\end{figure*}

\section{More Results}
\label{sup:more results}

\subsection{Qualitative Results on ScanNet}
Refer to~\cref{sec:exp_scannet}, {\ourmethod} outperforms two baseline methods on the ScanNet dataset when sparse observations are provided, which can be attributed to our proposed panorama inpainting technique that generates cross-view consistent panoramas, as described in~\cref{sec:multi-view diffusion}. We project generated meshes to panoramas in~\cref{fig:scannet_panorama} to demonstrate that the big portions of meshes are completed through our panorama inpainting.

\begin{figure*}[t]
\centering
\begin{subfigure}{\linewidth}
    \includegraphics[width=\linewidth]{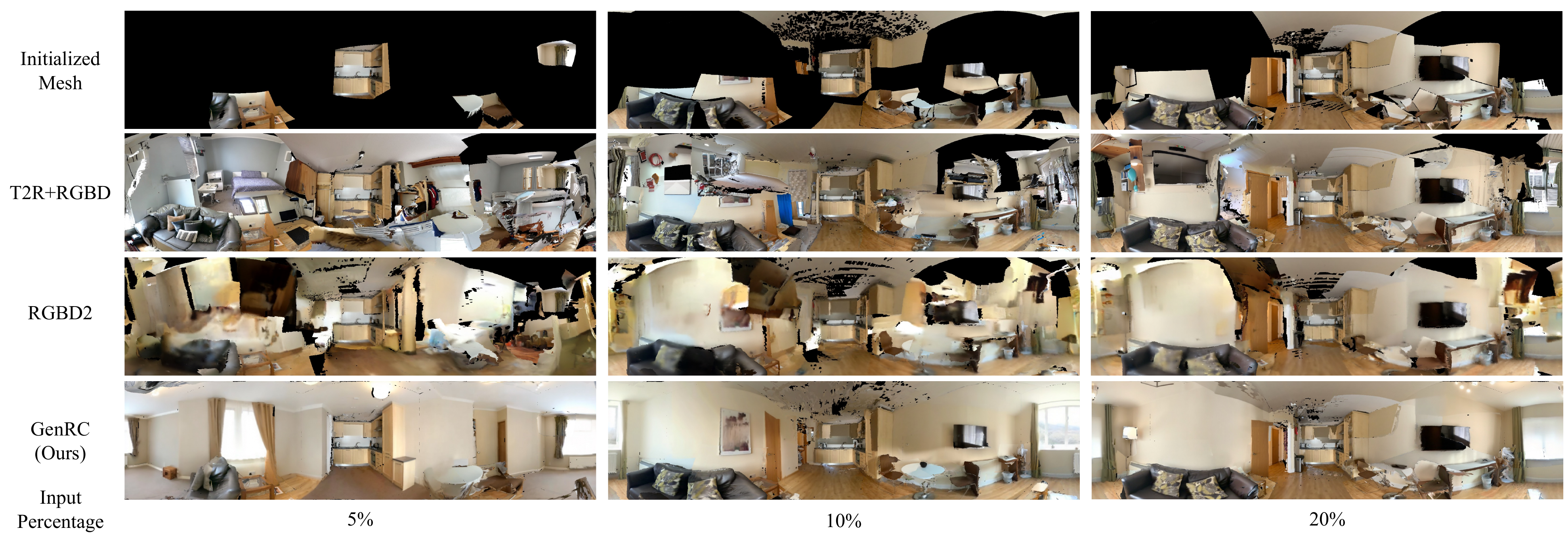}
    \caption{scene 41069021}
\end{subfigure}
\begin{subfigure}{\linewidth}
    \includegraphics[width=\linewidth]{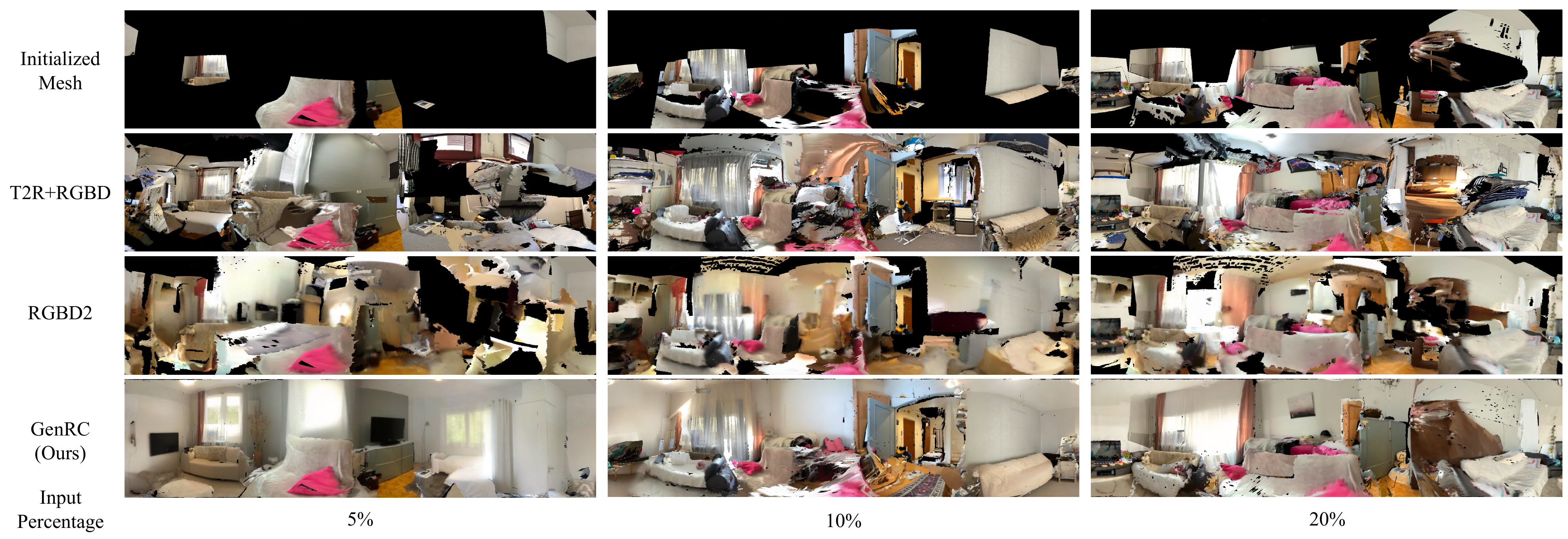}
    \caption{scene 47334360}
\end{subfigure}
\vspace{-0.5cm}
\caption{
    \textbf{Cross-domain results on ArkitScenes.} We project generated meshes to panoramas. When it comes to cross-domain data, RGBD2~\cite{lei2023rgbd2} and T2R+RGBD~\cite{hollein2023text2room} may generate unreasonable room structures, especially when the input observations are sparse (5\%). In contrast, {\ourmethod} can still generate cross-view consistent room meshes. Please refer to~\cref{sec:exp_arkits} for more quantitative discussions.
}
\label{fig:arkit_panorama}
\vspace{-0.5cm}
\end{figure*}
\begin{figure*}[t]
\centering
\includegraphics[width=0.9\linewidth]{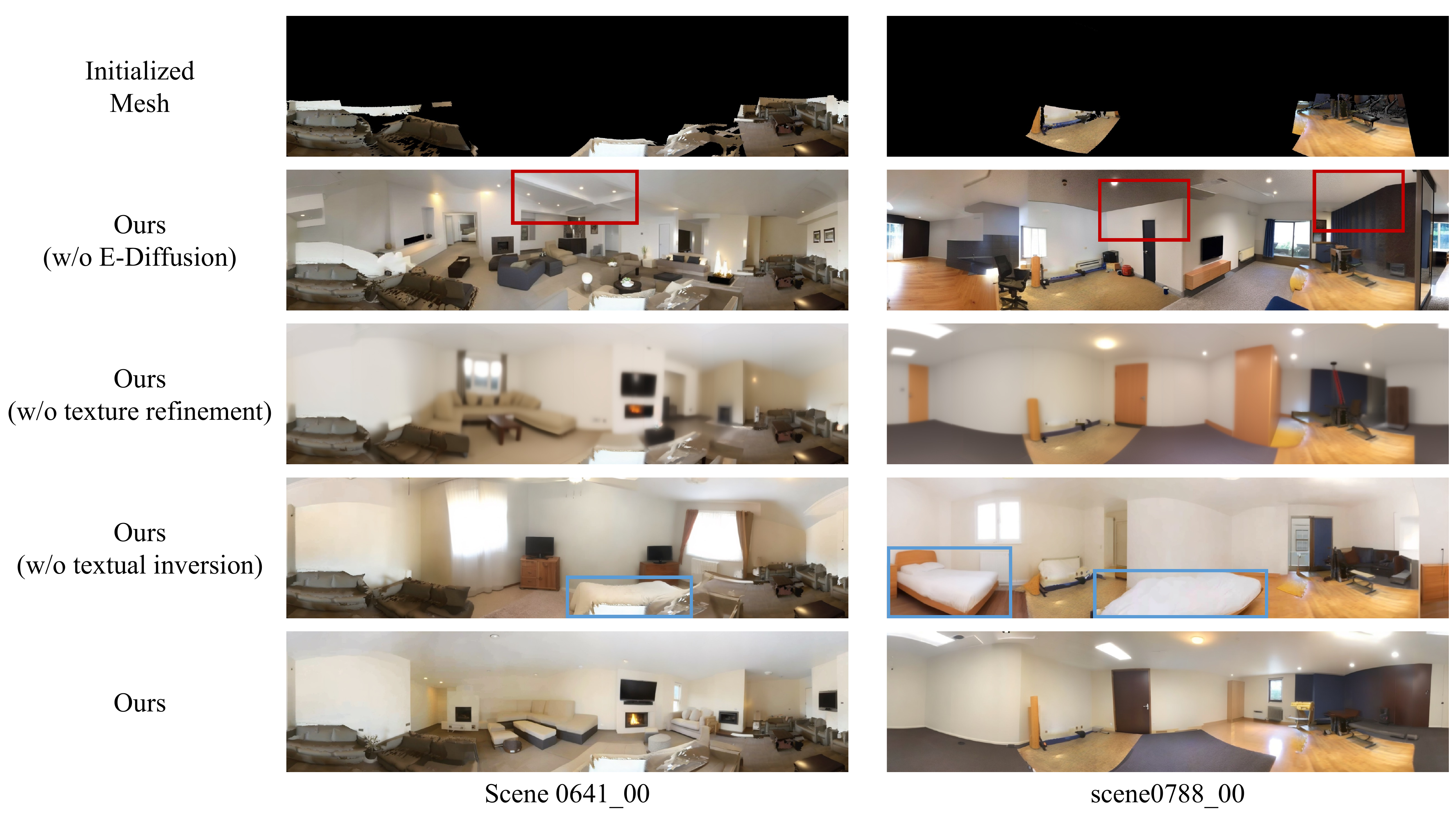}
\vspace{-0.36cm}
\caption{
    \textbf{Ablation studies on panorama inpainting and textual inversion.}     Directly applying MultiDiffusion~\cite{bar2023multidiffusion} for panorama inpainting (referred to as w/o E-Diffusion) can produce high-resolution panoramas. However, they don't satisfy the geometry of equirectangular projection (e.g., the edges between walls and ceilings appear straight in the red boxes). In addition, directly performing our proposed E-Diffusion without texture refinement causes blurry results. Without textual inversion, the Stable Diffusion~\cite{rombach2022high} model may generate objects (e.g., beds in the blue boxes) that are irrelevant to input images. Our method with all components can produce more detailed, stylistically coherent, and geometrically correct results. Please refer to~\cref{sec:ablation} for more quantitative discussions.
}
\label{fig:ablation_panorama}

\vspace{-0.5cm}
\end{figure*}
\begin{table}[t]
    \centering
    \caption{\textbf{Ablation studies on the number of views for E-Diffusion.} To inpaint a panorama with equirectangular geometry, E-Diffusion considers a set of overlapping views and denoise them concurrently (refer to~\cref{sec:multi-view diffusion}). We finally select 8 views in E-Diffusion because it results in the lowest depth means square error and highest CLIP score when sparse observations are given (i.e., 5\% and 10\%).
    }
    \resizebox{0.8\linewidth}{!} {
        \begin{tabular}[t]{l*{10}{c}}
             \toprule
               \multirow{2}{*}{\vspace{-0.2cm}Number of views} &
                \multicolumn{3}{c}{PSNR$_{\textrm{color}}$(↑)} & 
                \multicolumn{3}{c}{MSE$_{\textrm{depth}}$(↓)} &
                \multicolumn{3}{c}{CS$_{\textrm{input}}$(↑)} \\
             \cmidrule(lr){2-4}\cmidrule(lr){5-7}\cmidrule(lr){8-10}
                & 5\% & 10\% & 20\%  
                & 5\% & 10\% & 20\%
                & 5\% & 10\% & 20\%
                \\
             \cmidrule(lr){1-10}

                6 & 
                \textbf{14.5} & 16.4 & \textbf{17.6} & 
                0.29 & 0.16 & \textbf{0.09} &
                0.771 & 0.801 & \textbf{0.811}\\

                8 (Ours) & 
                14.4 & \textbf{16.7} & \textbf{17.6} & 
                \textbf{0.27} & \textbf{0.13} & \textbf{0.09} &
                \textbf{0.794} & \textbf{0.812} & 0.809\\

                16 & 
                \textbf{14.5} & \textbf{16.7} & \textbf{17.6} & 
                0.32 & 0.19 & 0.11 &
                0.772 & 0.801 & \textbf{0.811} \\

             \bottomrule
        \end{tabular}
    }
    \label{table:exp_ablation_EGMD_views}
    \vspace{-0.5cm}
\end{table}
\begin{table}[t]
    \centering
    \caption{\textbf{Ablation studies on the number of denoising steps for texture refinement.} 
    We utilize MultiDiffusion~\cite{bar2023multidiffusion} for the last $F$ denoising steps to refine the high-frequency texture (refer to~\cref{sec:multi-view diffusion}). Out of 50 denoising steps in the reverse diffusion process, we set the number of denoising steps for texture refinement as 20, which results in the lowest depth mean square error. Note that $F=50$ is equivalent to directly applying MultiDiffusion~\cite{bar2023multidiffusion} for panorama inpainting (as w/o E-Diffusion described in~\cref{sec:ablation}) and $F=0$ is equivalent to applying E-Diffusion without texture refinement.
    }
    \resizebox{0.8\linewidth}{!} {
        \begin{tabular}[t]{l*{10}{c}}
             \toprule
               \multirow{2}{*}{\vspace{-0.2cm}Number of Steps ($F$)} &
                \multicolumn{3}{c}{PSNR$_{\textrm{color}}$(↑)} & 
                \multicolumn{3}{c}{MSE$_{\textrm{depth}}$(↓)} &
                \multicolumn{3}{c}{CS$_{\textrm{input}}$(↑)} \\
             \cmidrule(lr){2-4}\cmidrule(lr){5-7}\cmidrule(lr){8-10}
                & 5\% & 10\% & 20\%  
                & 5\% & 10\% & 20\%
                & 5\% & 10\% & 20\%
                \\
             \cmidrule(lr){1-10}

                50 (w/o E-Diffusion) & 
                13.8 & 16.4 & 17.5 & 
                0.32 & 0.16 & \textbf{0.09} &
                0.765 & 0.799 & 0.808 \\

                40 & 
                14.4 & 16.6 & \textbf{17.7} & 
                0.31 & \textbf{0.13} & 0.10 &
                0.770 & 0.801 & \textbf{0.812} \\
                
                30 & 
                14.4 & 16.6 & 17.6 & 
                0.35 & 0.14 & 0.10 &
                0.772 & 0.801 & 0.811\\

                20 (Ours) & 
                14.4 & \textbf{16.7} & 17.6 & 
                \textbf{0.27} & \textbf{0.13} & \textbf{0.09} &
                \textbf{0.794} & \textbf{0.812} & 0.809\\

                10 & 
                14.3 & 16.6 & 17.5 & 
                0.37 & 0.15 & 0.12 &
                0.764 & 0.801 & 0.811 \\

                0 (w/o texture refinement) & 
                \textbf{14.6} & 16.4 & \textbf{17.7} & 
                0.35 & 0.21 & 0.13 &
                0.785 & 0.808 & 0.807\\
             \bottomrule
        \end{tabular}
    }
    \label{table:exp_ablation_texture_refinement}
\end{table}

\subsection{Qualitative Results on ArkitScenes}
Refer to~\cref{sec:exp_arkits}, {\ourmethod} demonstrates superior performance in both visual and geometric metrics on the ArkitScenes dataset. In~\cref{fig:arkit_panorama}, we can observe that when it comes to cross-domain data, RGBD2~\cite{lei2023rgbd2} and T2R+RGBD~\cite{hollein2023text2room} cannot successfully produce reasonable room structures when the input observation is sparse (i.e., 5\%). Refer to the 5\% results of RGBD2 and T2R+RGBD in~\cref{fig:arkit_panorama}, the 3D geometries are inconsistent along with unreasonable room structures that are not perpendicular to the ground. These point out the limitation of iterative methods which could still fail even given predefined trajectories. In contrast, {\ourmethod} can still generate visually pleasing room appearance and 3D consistent room structures even if the input observations are sparse and without predefined trajectories.



\subsection{Qualitative Results of Ablation Studies}
\label{sec:supple_qual}
{\ourmethod} generates high-fidelity and cross-view consistent panoramas via E-Diffusion, texture refinement, and textual inversion (refer to~\cref{sec:textual inversion} and~\cref{sec:multi-view diffusion}). We demonstrate the importance of these components by removing one of them at a time and the results are shown in~\cref{fig:ablation_panorama}.

\subsection{More Ablations}
\label{sec:more_abla}
We provide additional ablation studies on hyperparameter selection of E-Diffusion (refer to~\cref{sec:multi-view diffusion}): (1) the number of views used while E-Diffusion and (2) the number of denoising steps for texture refinement. As shown in~\cref{table:exp_ablation_EGMD_views} and ~\cref{table:exp_ablation_texture_refinement}, we consider 8 views in E-Diffusion and set the number of denoising steps for texture refinement as 20 out of 50 denoising steps in the reverse diffusion process.

\section{Future Works}


When a scene is cluttered with many objects, our approach may not complete the geometry of all 3D objects. We can further complete 3D objects through 3D object completion techniques such as~\cite{cheng2023sdfusion, kasten2023point, wu2023multiview, yan2022shapeformer} in future works.

\end{document}